\documentclass[sigconf]{acmart}

\AtBeginDocument{%
  \providecommand\BibTeX{{%
    \normalfont B\kern-0.5em{\scshape i\kern-0.25em b}\kern-0.8em\TeX}}}

\copyrightyear{2021}
\acmYear{2021}
\setcopyright{acmcopyright}
\acmConference[ASIA CCS '21]{Proceedings of the 2021 ACM Asia Conference on Computer and Communications Security}{June 7--11, 2021}{Hong Kong, Hong Kong}
\acmBooktitle{Proceedings of the 2021 ACM Asia Conference on Computer and Communications Security (ASIA CCS '21), June 7--11, 2021, Hong Kong, Hong Kong}
\acmPrice{15.00}
\acmDOI{10.1145/3433210.3437529}
\acmISBN{978-1-4503-8287-8/21/06}

\usepackage{graphicx}
\usepackage{subfig}
\usepackage{float}
\usepackage{comment}
\usepackage{color}
\usepackage{hhline}
\usepackage{booktabs}
\usepackage{multirow}
\usepackage{xspace}
\usepackage{colortbl}
\usepackage{array}
\usepackage{epsfig}

\usepackage{pifont}
\newcommand{\blackding}[1]{\ding{\numexpr181+#1\relax}}

\newcommand{\af}{\vspace*{-3pt}}

\definecolor{mygray}{gray}{.8}

\newcommand{\oursys}{\textsc{Themis}\xspace}

\newcommand{\zedit}[1]{\textcolor{black}{#1}}
\newcommand{\fedit}[1]{\textcolor{black}{#1}}

\begin{document}
\fancyhead{}
\title{Exploiting the Sensitivity of $L_2$ Adversarial Examples \\ to \emph{Erase-and-Restore}}


\author{Fei Zuo}
\affiliation{%
  \institution{University of South Carolina}
  \city{Columbia, SC}
  \country{USA}}
\email{fzuo@email.sc.edu}

\author{Qiang Zeng}
\affiliation{%
  \institution{University of South Carolina}
  \city{Columbia, SC}
  \country{USA}}
\email{zeng1@cse.sc.edu}


\begin{abstract}

By adding carefully crafted perturbations to input images, adversarial examples (AEs) can be generated to mislead neural-network-based image classifiers. 
   $L_2$ adversarial perturbations by Carlini and Wagner (CW)
   are among the most effective but difficult-to-detect attacks.
   While many countermeasures against AEs have been proposed, 
   detection of adaptive CW-$L_2$ AEs is still an open question.
   We find that,
   by randomly erasing some pixels in an $L_2$ AE and then restoring it with an inpainting technique, the AE, before and after the steps, tends to have different classification results, while a benign sample does \emph{not} show this symptom. 
   We thus propose a novel AE detection technique, Erase-and-Restore (E\&R), that exploits the intriguing sensitivity of $L_2$ attacks. 
   Experiments conducted on two popular image datasets, CIFAR-10 and ImageNet, show that the proposed technique is able to detect over 98\% of $L_2$ AEs  and has a very low false positive rate on benign images. The detection technique exhibits high transferability: a detection system trained using CW-$L_2$ AEs can accurately  detect AEs generated using another $L_2$ attack method.
   More importantly, our approach demonstrates strong resilience to adaptive $L_2$ attacks, filling a critical gap
   in AE detection. 
   Finally, we interpret the detection technique through both visualization and quantification.

\end{abstract}




\maketitle
\pagestyle{empty} 

\section{Introduction}
By adding deliberately crafted perturbations into an image,
an attacker is able to create an \emph{adversarial example} (AE), which misleads a neural-network-based classifier to output an incorrect prediction result. Worse, the malicious perturbations in an AE are so subtle that they are usually human-imperceptible. 
As neural networks are increasingly deployed, AEs raise crucial security concerns especially in many vision-related applications. 

The term \emph{adversarial example} can be formally defined as follows. For a pre-trained DNN $f$, let $x$ be an original image. An adversarial example $x^{adv}$, derived from $x$, can guide the model $f$ to make an incorrect prediction. Moreover, to hide the adversarial perturbation, the generation of $x^{adv}$ is equivalent to solve the following constrained optimization problem:
\begin{equation} \label{equation:adv}
\begin{aligned}
 &\min\limits_{x^{adv}}\ \Vert x^{adv} - x\Vert_p \\
 &\mathrm{s.t.}\ \ {y}'= f(x^{adv}),\ 
y =  f(x),\ \text{and}\ y \neq {y}'
 \end{aligned}
\end{equation}
where $y$ and ${y}'$ are respectively the prediction results of feeding $x$ and $x^{adv}$ to $f$. 

\begin{figure} 
\centering
\subfloat[Original image]{\includegraphics[scale=.33]{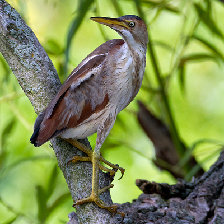}}
\ \ 
\subfloat[Corrupted image]{\includegraphics[scale=.33]{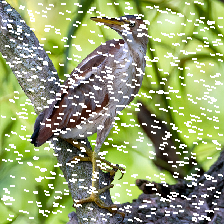}}
\ \ 
\subfloat[Restored image]{\includegraphics[scale=.33]{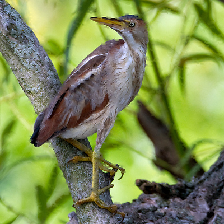}}
\caption{Restoring lost parts of an image with inpainting.}\label{fig_inpaint}
\end{figure}

To gauge such adversarial perturbations, $L_p$ norms are usually used to quantitatively describe the discrepancy between  $x$ and $x^{adv}$. According to the value of $p$ in Equation~\ref{equation:adv}, the mainstream AE generation algorithms can be categorized
into three families: $L_0$, $L_2$ and $L_\infty$ attacks. Informally, $L_0$ measures the number of modified pixels, $L_2$ the Euclidean distance between $x$ and $x^{adv}$, and $L_\infty$ the largest modification among all the modified pixels. 

As suggested by Carlini and Wagner~\cite{carlini2017towards}, defenders should consider evaluating  ``\textit{a powerful attack}'' and particularly  emphasized $L_2$ attacks (Section 9 in~\cite{carlini2017towards}). Other researchers also agree that $L_2$ attacks by Carlini and Wagner (CW)~\cite{carlini2017towards} 
``\textit{are among the most effective white-box attacks and should be used among 
the primary attacks to evaluate potential defences}''\cite{art2018}. 
Although researchers have proposed many AE detection methods~\cite{li2017adversarial,metzen2017detecting,meng2017magnet,xu2017feature}, recent studies~\cite{he2017adversarial,carlini2017magnet,carlini2017adversarial} show that 
the detection usually goes ineffective when facing 
adaptive CW-$L_2$ AEs. 
Thus, how to accurately detect adaptive $L_2$ AEs is still an open question.
We focus on tackling $L_2$ AEs in this work, and  
our goal is a technique that not only detects $L_2$ AEs accurately but is also resilient to adaptive attacks.

We have two key insights. First, we observe that 
those deliberately  corrupted pixels exert a malicious influence \emph{altogether} (e.g., through multiple rounds of optimizations during AE generation). 
It implies that a destruction of the completeness of the influence by the perturbed pixels
can cause a failure of the attack. Second, while destruction may also harm
the classification accuracy for benign samples,
there exist very effective
\emph{inpainting} techniques~\cite{shen2002mathematical,telea2004image,mairal2007sparse} in the image processing area that can help restore a partially corrupted image. For example, Figure~\ref{fig_inpaint}(a) shows an original image, and Figure~\ref{fig_inpaint}(b) a corresponding  corrupted image where many regions are erased. After inpainting, as shown in Figure~\ref{fig_inpaint}(c), the corrupted image is well restored.

Thus, we hypothesize that if we \textbf{\emph{randomly}} erase a portion of  pixels from an AE and then 
apply inpainting to it, the attack will probably fail for two reasons. 
Discarding many small regions from an AE will ruin the holistic adversarial influence formed 
by the maliciously perturbed pixels. Second, the inpainting typically restores the image in a benign way that does \emph{not} preserve the malicious influence. 
By contrast, if we apply the same ``\emph{Erase-and-Restore}'' (E\&R) operations to a benign
sample, the classification results, before and after the steps, tend to be similar, as inpainting 
by design is to reverse deterioration of benign images.

\begin{figure*} [!thb]
\centering

\subfloat[Adversarial examples]{\includegraphics[scale=0.4]{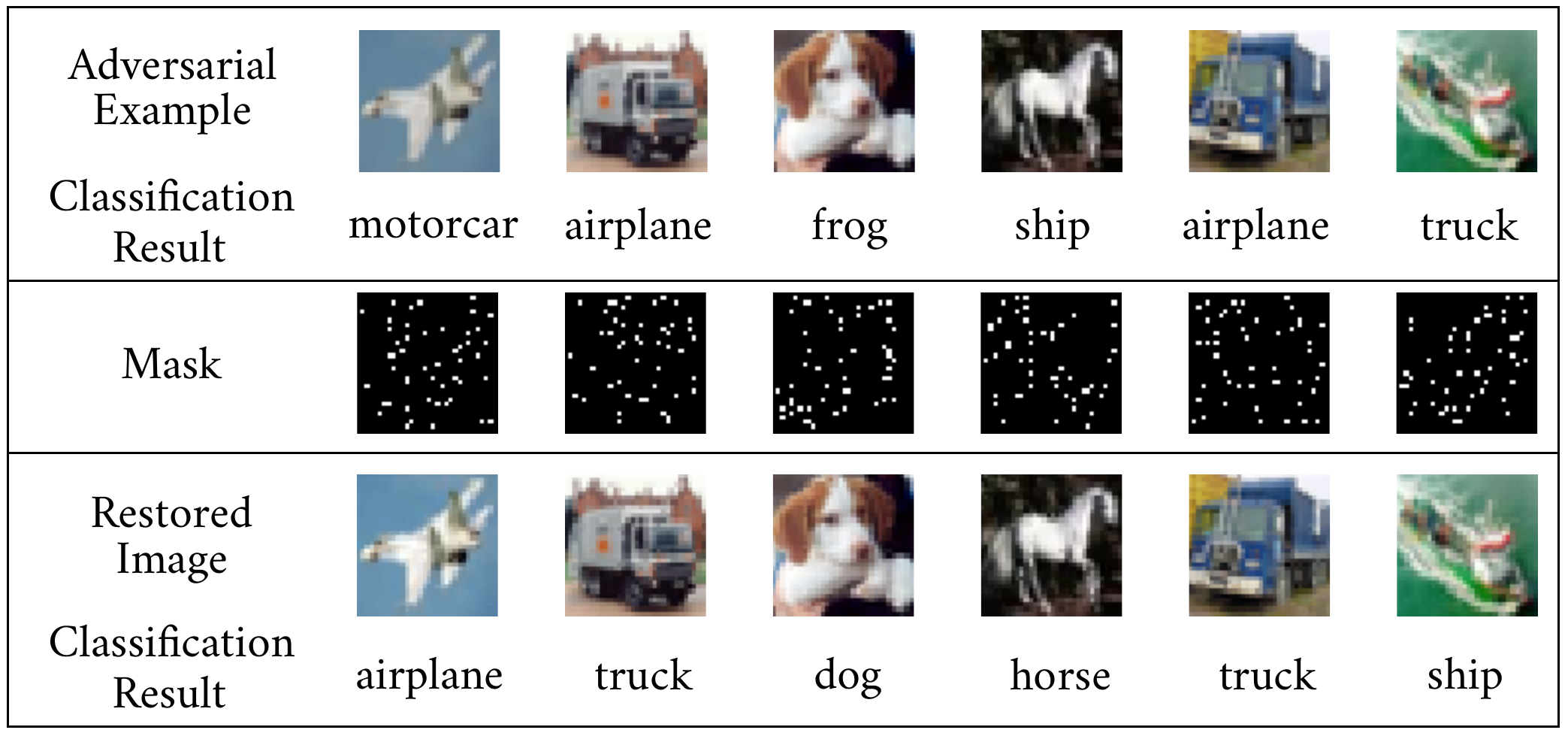}}
\quad
\subfloat[Benign samples]{\includegraphics[scale=0.4]{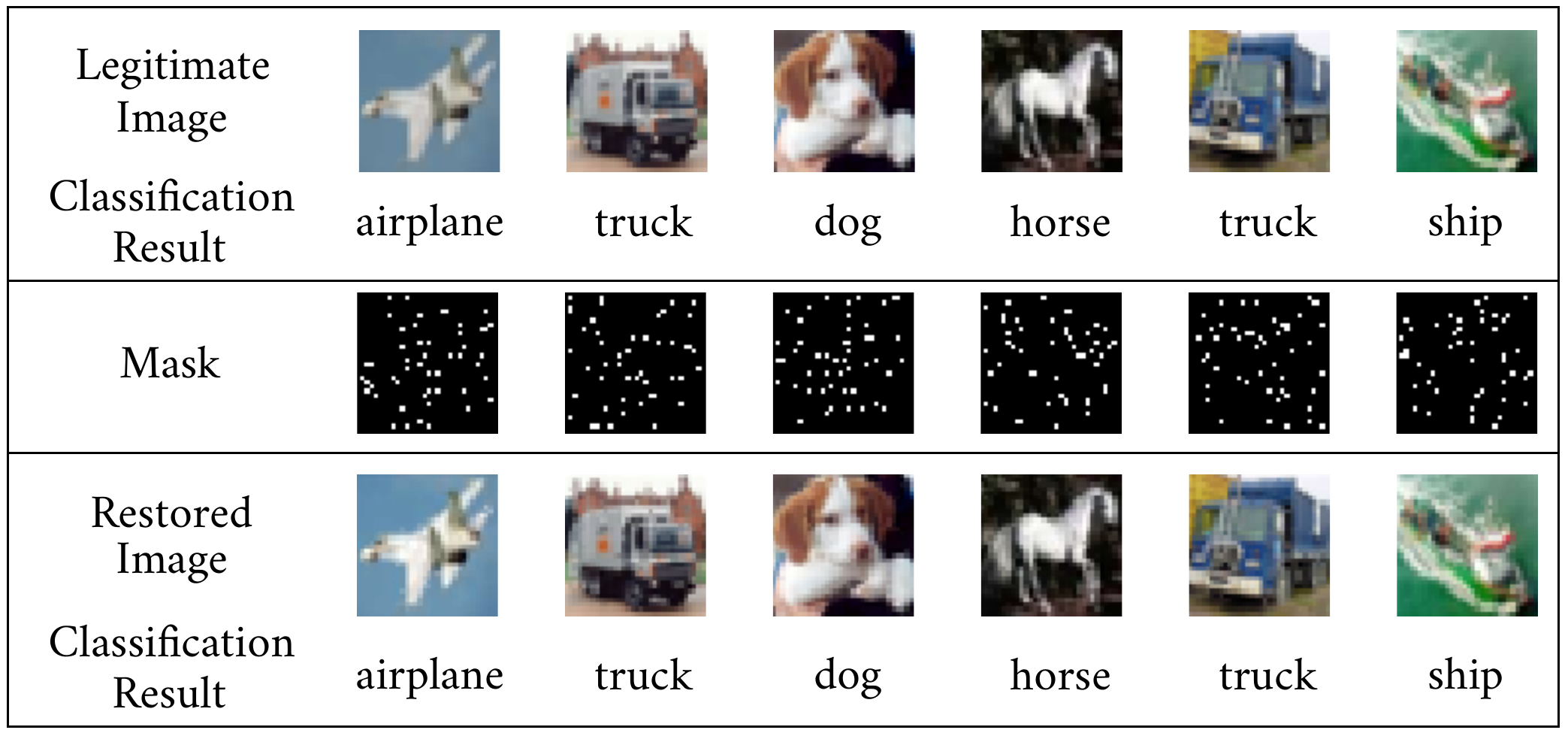}}
\caption{Different impacts of  ``Erase-and-Restore'' on AEs and benign samples.}\label{fig:adv_ben}
\end{figure*}

Figure~\ref{fig:adv_ben} illustrates our insights and observations using six color images from CIFAR-10. 
A \emph{random mask} (mask, for short) in our work describes the locations of pixels that
are randomly erased. We randomly erase 5\% of the pixels of each image. 
The AEs are generated using the CW algorithm~\cite{carlini2017towards}.
As shown in Figure~\ref{fig:adv_ben}(a), 
the classification results of each AE, before and after 
the E\&R operations, are different. 
By contrast, as shown in Figure~\ref{fig:adv_ben}(b), the classification results of each
benign sample, before and after the steps, are the same. 
Our large-scale experiments (Section~\ref{sec:sys}) also show consistent results. 

We consider the sensitivity to E\&R operations as an exploitable characteristic of $L_2$ AEs, 
and propose a novel AE detection technique:
given an image, if the classification results before and after E\&R
vary greatly, it is an AE; otherwise, a benign sample.
We accordingly implement an  $L_2$ AE detector, named \oursys. To improve the detection accuracy,
it is enhanced by applying E\&R multiple times. 
Specifically, given an image $I_0$, we 
\emph{randomly} erase some pixels of $I_0$ each time to create a sequence of images $\{I_1, I_2, \cdots, I_n\}$. 
Next, an inpainting technique is applied to them to obtain the restored images $\{I'_1, I'_2, \cdots, I'_n\}$. Finally, a classifier makes use of the prediction results of $I_0$ and the restored images to determine whether $I_0$ is an AE.

We have evaluated our system using the popular image datasets
CIFAR-10 and ImageNet. Two widely-discussed $L_2$ AE generation methods, CW~\cite{carlini2017towards} and DeepFool~\cite{moosavi2016deepfool}, are considered in the evaluation.
Our experiments show that the proposed detection technique is very effective. Take the CW~\cite{carlini2017towards} attack as an example, on the CIFAR-10 dataset, \oursys can detect 100\% AEs with a false positive rate (FPR)=0,
and on ImageNet, it can detect 99.3\% AEs with FPR = 2.7\%. 
In addition, the detection technique demonstrates three notable characteristics.
\blackding{1} It is \textbf{target-model agnostic}: a detector trained using AEs targeting one neural network model
can be directly used to detect AEs targeting another. \blackding{2} It has good \textbf{transferability}: 
a detector trained using AEs generated by one attack method can be directly used to
detect AEs by another. \blackding{3} More importantly, it shows \textbf{high resilience to adaptive attacks}. Finally, we interpret the effectiveness of 
the detection technique through both visualization and quantification. 

The key contributions of our work include:

\begin{itemize}

\item We find an interesting characteristic of $L_2$ AEs, whose classification 
results vary sharply when Erase-and-Restore operations are applied; 
meanwhile, benign samples are not so sensitive. 

\item We propose to exploit
the characteristic for AE detection, and employ the idea of sampling to enhance 
the detection. By applying E\&R for multiple times, 
richer features are generated to improve the detection accuracy.

\item We implement the detection technique in \oursys and evaluate it
on two popular datasets, CIFAR-10 and ImageNet. The experiment results show that \oursys outperforms prior techniques (such as NIC~\cite{ma2019nic}, LID~\cite{ma2018characterizing}, and Feature Squeezing~\cite{xu2017feature}), achieving not only
\textbf{the highest detection rate} but also the \textbf{lowest false positive
rate}. We are to make the source code, 
datasets, and models of this work publicly available.\footnote{\url{https://github.com/quz105/Erase-and-Restore}.}

\item The detection technique is target-model agnostic and shows high transferability across different $L_2$ attack methods. Furthermore, it demonstrates strong resilience to adaptive CW-$L_2$ attacks, filling
a critical gap in AE detection.

\item We interpret the effectiveness of 
the detection technique in multiple ways.

\end{itemize}

\section{Background and Threat Model}\label{sec:bck}

\subsection{Attack Algorithms}

Adversarial attacks can be 
categorized as either non-targeted or targeted ones. The aim of
a non-targeted attack is to make the input be classified as any
arbitrary class except the correct one. By contrast, the aim
of a targeted attack is a specific attacker-desired incorrect
result, which is more threatening.
Next, we briefly describe the two most popular $L_2$ AE generation methods.

\vspace{3pt}
\noindent \textbf{Carlini \& Wagner Attacks}
Carlini and Wagner~\cite{carlini2017towards} designed a group of targeted AE generation methods which are denoted as CW attacks. According to the distance metrics adopted in an optimization target, CW attacks can be divided into three types: $L_0$-, $L_2$- and $L_{\infty}$-norm. In this paper, we mainly examine \emph{CW-$L_2$ attacks}, which are the most difficult to detect~\cite{he2017adversarial,carlini2017adversarial}.  

Due to a few creative designs, the CW attacks achieve  performance superior to other attack methods.
The first and foremost innovative design is using a logits-based objective function rather than softmax-cross-entropy loss, which plays a key role in the resilience improvement of the attack against 
defensive distillation~\cite{papernot2015distillation}.
Secondly, this algorithm maps the target variable to a space of the inverse trigonometric function, so that the problem is suitable to be solved by a modern optimizer, e.g. Adam~\cite{kingma2014adam}. Finally, a \emph{confidence-level} parameter $\kappa$ 
is introduced; as $\kappa$ increases, the model classifies the resulting AE as the attacker-desired label more likely,  giving the attacker
flexibility to make a trade-off between the degree of perturbations and misclassification probability.

\begin{figure*} [!thb]
\centering

\subfloat[Benign samples]{\includegraphics[scale=0.5]{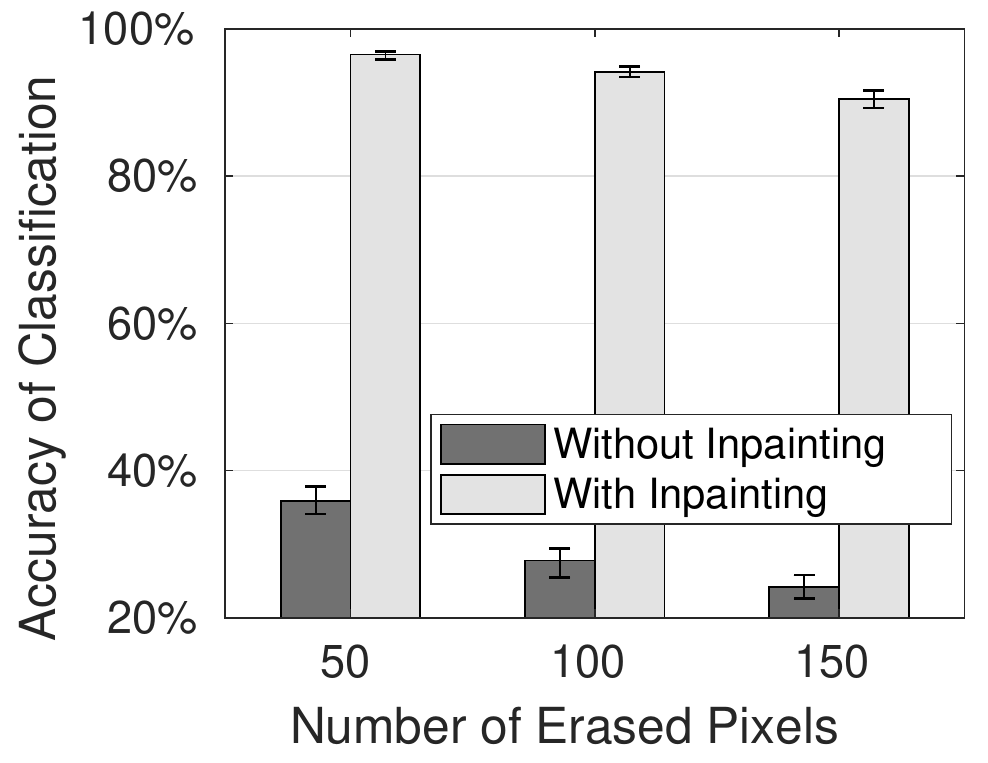}}
\ \ 
\subfloat[Success rates of AEs with E\&R]{\includegraphics[scale=0.5]{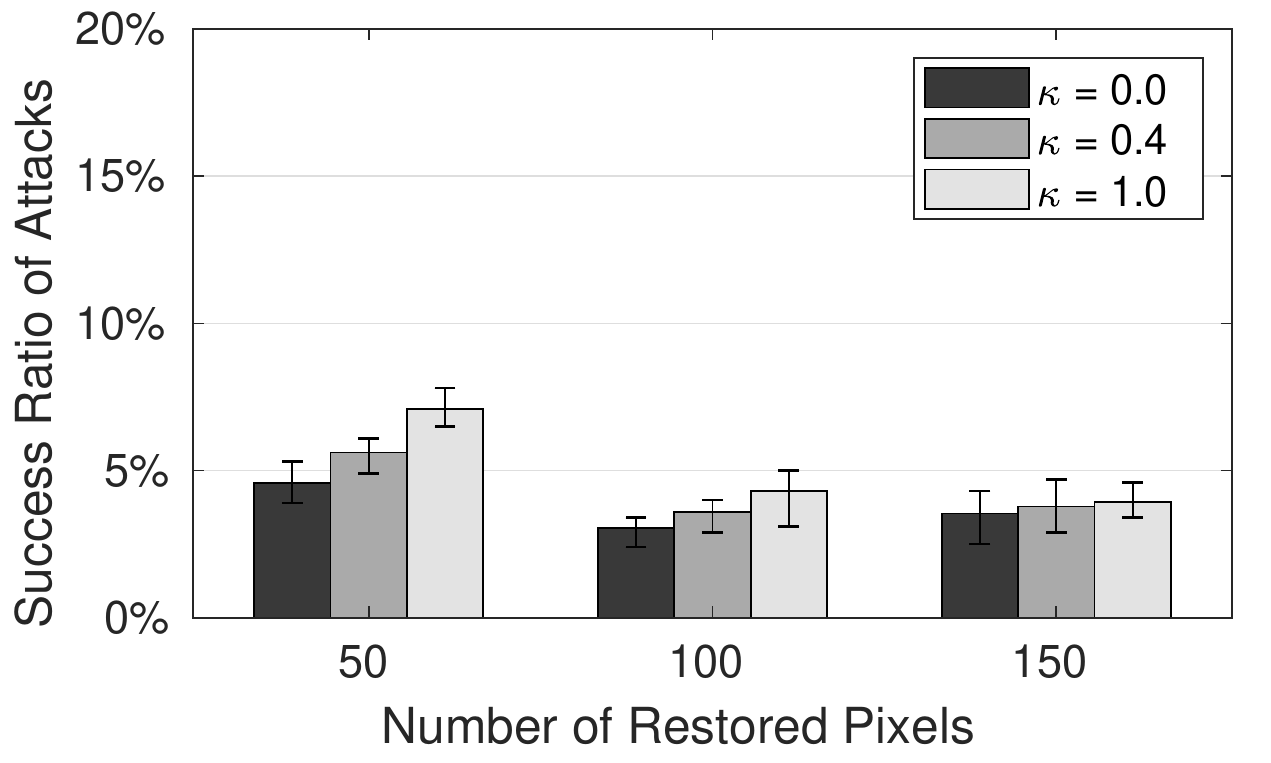}}
\caption{Impacts of E\&R on benign samples and AEs.}\label{fig_sensitive}
\end{figure*}

\vspace{3pt}
\noindent \textbf{DeepFool} \label{sec:deepfool}
Moosavi et al.~\cite{moosavi2016deepfool} developed the DeepFool attack that is used to create non-targeted AEs. The algorithm utilizes an iterative linearization of the classifier to generate $L_2$ minimization-based perturbations. To simplify the problem, the neural networks are imagined to be linear, so that the decision boundaries are a set of hyper-planes. Consequently, a polyhedron can be used to describe the output space. Assuming that $f$ is a binary differentiable classifier, to mislead the decision of $f$ near the current point $x_i$, the minimal perturbation is the orthogonal projection of $x_i$ onto the separating hyper-plane. At each iteration the minimal perturbation of the linearized classifier is computed as
\begin{equation}
    \mathrm{arg}\min\limits_{\delta_i}\|\delta_i\|_2\quad \mathrm{s.t.}\  f(x_i)+\nabla f(x_i)^T{\delta_i}=0
\end{equation}
where $\delta_i$ is the perturbation imposed on $x_i$. Note that neural networks are not actually linear, so the search is repeated until a successful AE is found. 

\subsection{Threat Model}

The adversary has full knowledge of the target model (including both its architecture and parameters). He also knows the existence and internal details of the detector, and is allowed to \emph{adapt attacks}. In adaptive attacks, the attacker tries to fool the image classifier and the detector at the same time.
We consider adaptive attacks and evaluate the resilience of our
detector to them in Section~\ref{sec:adapt}.

\section{Experimental Setup}\label{sec:setup}

Before presenting our defense scheme, we introduce the image datasets and the corresponding target neural networks on which we verify our key insights and evaluate the proposed approach.

\vspace{3pt}
\noindent \textbf{Image datasets.} We generate AEs using two popular datasets: CIFAR-10 and ImageNet, both of which are widely used in image classification tasks. In particular, for ImageNet, we adopt the \emph{ILSVRC2012}
samples to keep consistent with the prior state-of-the-art AE detector~\cite{ma2019nic}. 

\vspace{3pt}
\noindent \textbf{Target neural network models.}
(1) For CIFAR-10, we use two neural networks as the target models: a 32-layered ResNet model~\cite{he2016deep} (denoted as \emph{ResNet32}), and 
a model structure described in~\cite{carlini2017towards} (denoted as \emph{Carlini}). We train these two target neural network models from scratch (the accuracies of the two models are 91.96\% and 78.86\%, comparable with those published in prior works~\cite{ma2019nic,xu2017feature}). (2) For ImageNet we re-use a 50-layered ResNet model~\cite{he2016deep} provided in Keras~\cite{chollet2015keras} (denoted as \emph{ResNet50}). 

\vspace{3pt}
\noindent \textbf{AE generation and data preparation.}
Like existing AE detection works, only images that are correctly classified
by the corresponding target model are used to generate AEs in our experiments. To generate \emph{targeted} AEs, we designate the \textit{next} class as the target 
class, similar to many other AE detection works~\cite{ma2019nic,xu2017feature,zuo2019l0}.  
Only AEs that can successfully fool the target models
are used in the evaluation. For ImageNet, we collect 30,000 legitimate images
and create 30,000 AEs: DeepFool and CW-$L_2$ generate 15,000 AEs each. 
The number of  CW-$L_2$
AEs with each given confidence level (i.e.,$\kappa$= 0.0, 0.4, and 1.0) is the same, that is 5,000 for each sub-group. In the dataset, 80\% of instances are used for training and the remaining 20\% for testing, denoted as $\mathcal{D}_I$-\texttt{Train} and $\mathcal{D}_I$-\texttt{Test}, respectively. Similarly, for CIFAR-10, based on the types of target model, we have four dis-joint datasets, $\mathcal{D}_C$-\texttt{Carlini}-\texttt{Train}, $\mathcal{D}_C$-\texttt{Carlini}-\texttt{Test}, $\mathcal{D}_C$-\texttt{ResNet}-\texttt{Train}, and $\mathcal{D}_C$-\texttt{ResNet}-\texttt{Test}. The former two and the latter two datasets have the same size and data composition as $\mathcal{D}_I$-\texttt{Train} and $\mathcal{D}_I$-\texttt{Test}, respectively. All AEs are generated using the opensource tool \texttt{Foolbox}~\cite{rauber2017foolbox}. 

\vspace{3pt}
\noindent \textbf{Inpainting algorithm.}
The inpainting algorithm we choose in this work is designed by Telea~\cite{telea2004image}.
This inpainting algorithm needs to solve an \emph{Eikonal} equation, which is rarely differentiable everywhere. 
Considering the inpainting algorithm  
is \emph{not} fully differentiable, it results in a non-negligible obstacle for adaptive attackers.

\vspace{3pt}
The experiments were performed on a computer running
the Ubuntu 18.04 operating system with a 64-bit 3.6 GHz Intel\textsuperscript{\textregistered} Core\textsuperscript{(TM)} i7 CPU, 16 GB RAM and a GeForce\textsuperscript{\textregistered} GTX 1070 GPU.

\section{The Proposed Approach}\label{sec:sys}

\subsection{Our Insights}

\noindent \textbf{Effects of erasing (or adding noises) alone.}
Due to the optimization nature of AE generation methods like CW and DeepFool,
maliciously manipulated pixels in an AE are deliberately selected and perturbed.
Thus, each of the perturbed pixels plays a certain role in the attack. 
By \emph{randomly} \emph{erasing} many pixels of an input image, it is likely to
corrupt some of the perturbed pixels or their surrounding pixels 
in an AE, rendering the attack ineffective. 

In the case of \emph{benign} samples, however, the erasing operation, which is equivalent 
to introducing random noises to images, 
will significantly degrade the accuracy of the classifier.
The close correlation between the image quality and the accuracy of image classification 
has been widely studied in previous works~\cite{diamond2017dirty,da2016empirical,dodge2016understanding}. 
They mention that neural networks are
susceptible to random noise distortions. For example,
Costa et al.~\cite{da2016empirical} point out that  
\textit{``noises can hinder classification performance considerably and make classes harder to separate}.''

\vspace{3pt}
\noindent \textbf{Combining erasing and inpainting.} 
We thus propose to apply \emph{inpainting} after the erasing operation.
Inpainting is a category
of techniques for restoring damaged regions of images.
Given an erased region, an inpainting technique  
infers and recovers its original pixels. \emph{Our insight} is that, while inpainting works very well
for recovering benign samples, its recovering effect is usually \emph{not}
what the AE attacker desires, as the maliciously perturbed
regions, once erased, can hardly be recovered to the attacker-intended values. 

We further design experiments to verify the two insights in Section~\ref{sec:verify}.

\vspace{-2pt}
\subsection{Verifying Our Insights} \label{sec:verify}

From CIFAR-10, we randomly select 1,000 images that can be correctly 
classified by \emph{ResNet32}. 
As shown in Figure~\ref{fig_sensitive}(a), after randomly erasing 50$\sim$150 (around 5\%$\sim$15\%) of the pixels in each image,
without inpainting, the classification accuracy significantly degrades from 100\% to the range from 24.2\% (when erasing 15\%) to 35.9\% (when erasing 5\%), which verifies that erasing alone harms
the classification accuracy for benign images significantly. 
By contrast, with inpainting applied, the classification accuracy recovers to 90.5\%$\sim$96.6\%. 

Besides, for each benign image we use the CW algorithm to generate three AEs  with three 
different confidence levels ($\kappa=$ 0.0, 0.4, and 1.0, respectively). 
All the AEs successfully 
fool the \emph{ResNet32} model. As shown in Figure~\ref{fig_sensitive}(b), after 
randomly erasing  50$\sim$150 (around 5\%$\sim$15\%) of the pixels in each AE and then restoring them using inpainting, the success rate of attacks dramatically decreases from the original 100\% to the range 3.1\%$\sim$7.1\%. 

Similar results can be observed on the ImageNet dataset as well. (1) Specifically, we randomly select
1,000 images from ImageNet that can be correctly classified by the \emph{ResNet50} model. For example,
after erasing and restoring 5\% of the pixels in each image, the classification accuracy 
stays at 96.3\%. (2) On the other hand, when we apply the same erasing and restoring operations to 
the 1,000  AEs generated from 
these benign images, the success rate of attacks decreases from 100\% to around 4.1\%. 

Therefore,  it can be concluded that E\&R has very small impacts on benign samples, but large impacts on AEs, demonstrating a noticeable contrast. 

\subsection{Approach Details} \label{sec:details}

\begin{figure*} 
\centering
\includegraphics[scale=0.45]{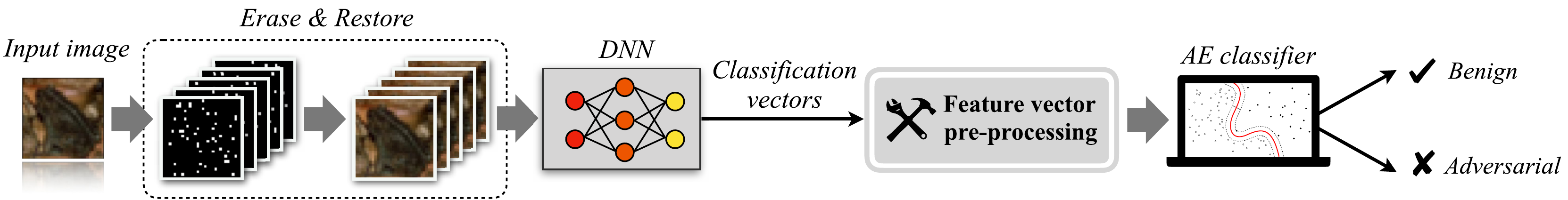}
\caption{Architecture of \oursys.}\label{design}
\end{figure*}

Based on our insights, we propose a novel AE detection technique, named E\&R, that
exploits the sensitivity of AEs to E\&R operations, and implement it in a system, called \oursys, as shown in  Figure \ref{design}.
(1) Given an input image $I_0$, we \textbf{\em randomly erase $\lambda$ pixels} of it to create a deteriorated image $I$. Employing the idea of 
sampling, this step
is repeated for $n$ times to obtain a sequence of deteriorated images $\{I_1, I_2, \cdots, I_n\}$. The \emph{intuition} behind it is that even if an AE ``luckily'' evades 
the detection once, it is very unlikely for it to hide itself throughout the multiple samples. 
(2) Next, an inpainting technique is leveraged to produce a corresponding sequence of \textbf{\em restored} images $\{I'_1, I'_2, \cdots, I'_n\}$. (3) Finally, we feed both the input image $I_0$ and $\{I'_1, I'_2, \cdots, I'_n\}$ into a neural-network classifier, and collect all the classification results. 

Given an image in CIFAR-10, its classification result is a vector $\in\mathbb{R}^{10}$ (since there are 10 classes in the dataset).
We simply concatenate all the classification-result vectors for both $I_0$ and $\{I'_1, I'_2, \cdots, I'_n\}$
to obtain a feature vector $\in\mathbb{R}^{10\times(n+1)}$ for training
the AE classifier.

\begin{table*}[!tb]
\centering
\caption{Performance of \oursys. After \oursys is trained using training datasets that contain benign samples, CW and DeepFool AEs, \fedit{the detection rate and FPR (the rate of benign samples misclassified as AEs) are measured using testing sets.}}\label{tab:cw} 

\begin{tabular}{c|c||c|c|p{1.2cm}<{\centering}|p{1.2cm}<{\centering}|p{1.2cm}<{\centering}|c}
\specialrule{.1em}{.05em}{.05em}
      \multirow{2}{*}{\textbf{Dataset}} & \multirow{1}{*}{\textbf{Target}} &  \multirow{2}{*}{\textbf{Classifier}} &
      \multirow{2}{*}{\textbf{FPR}} & \multicolumn{3}{c|}{\textbf{Detection Rate: CW-$L_2$}} & \multirow{1}{*}{\textbf{Detection Rate:}} \\ \cline{5-7}
      
    & \textbf{Model} & & & $\kappa$=0.0 & $\kappa$=0.4 & $\kappa$=1.0 & \textbf{DeepFool}\\ \specialrule{.1em}{.05em}{.05em}
\multirow{4}{*}{CIFAR-10} &  \multirow{2}{*}{Carlini}  & SVM & {0.6}\% & 100\% & {100\%} & {100\%} & {99.4}\% \\ \cline{3-8} 
&  & AdaBoost & 0.0\% & 100\%   & 100\% & 100\% & 98.3\% \\ \hhline{~|-|-|-|-|-|-|-}
                          &  \multirow{2}{*}{ResNet32}    & SVM & {2.8\%} & {99.4\%}   & {99.6\%} & {99.6\%}&{99.8\%} \\ \cline{3-8}
                          &  & AdaBoost & 0.9\% & 99.4\%  & 99.2\% & 99.4\% & 99.8\%\\  
                          \hline
\multirow{2}{*}{ImageNet}    & \multirow{2}{*}{ResNet50} & SVM &  {3.5\%} & {97.9\%} & {98.4\%} & {98.7\%} & 93.7\% \\ 
 \cline{3-8} 
&  & AdaBoost & 2.7\% & 98.9\%   & 99.2\%  & 99.3\% & 95.0\%\\
\specialrule{.1em}{.05em}{.05em}
\end{tabular}
\end{table*}

Given an image from the ImageNet, its classification result is a vector $\in\mathbb{R}^{1000}$ (since there are 1,000 classes in the dataset). 
Thus, the number of features to be fed to our classifier is $1000\times(n+1)$, which is too large.
To make the training of our classifier more feasible,
Principal Component Analysis (PCA) is performed on the classification results 
 of $I_0$ and $\{I'_1, I'_2, \cdots, I'_n\}$, to reduce the dimensionality to a lower value $d$. 
Unless otherwise specified, we set $d$ to 10 (1\% of the original dimensionality) to keep consistent with CIFAR-10. Note that the number of principal components should be less than both the number of features and the number of samples,
when solving PCA based on the truncated SVD (singular value decomposition). In our case, the number of samples is $n+1$;
we thus let $n=11$ (we discuss the impact of $n$'s values with detailed experimental results in Section~\ref{sec:n}). We concatenate the vectors of principal components for
both $I_0$ and $\{I'_1, I'_2, \cdots, I'_n\}$ to obtain a feature vector for training our classifier. 

The value of the parameter $\lambda$ (number of pixels to be erased) is set to 10\% of the pixels in an input image.
We adopt this value for two reasons. (1) As shown in Figure~\ref{fig_sensitive}, when 10\% of
the pixels are erased and restored, it harms the success rate of AEs most heavily, without 
degrading the classification accuracy for benign samples significantly. (2) The 
inpainting algorithm we adopt performs very well when
the portion of corrupted pixels in an image is less than 15\%~\cite{telea2004image}. 

It is worth mentioning that $\lambda=10\%$  leads to an enormous randomness pool. 
Take an image in CIFAR-10 as an example, the size of which is 32$\times$32:
with $\lambda$=100 ($\approx10\%$ of the pixels),
the number of unique masks is around 7.7$\times 10^{140}$. 
It is thus very unlikely for an adaptive attacker to correctly predict which masks will be used
by our detector.

We train our AE classifier using two supervised learning techniques: \emph{AdaBoost}~\cite{freund1997decision} and \emph{SVM}~\cite{cortes1995support}.

\section{Evaluation}\label{sec:eval}

We evaluate the detection performance of the proposed
scheme against $L_2$ attacks in terms of \textit{detection rate} and \emph{false positive rate} (FPR). 
The detection rate is defined as the ratio of the number of successfully detected
AEs to the total number of AEs. 
FPR refers to the fraction of benign samples
that are misclassified as AEs.

\subsection{Detection Performance}

\begin{table*}[!tb]
\centering
\caption{Comparison with other AE detectors (DR: Detection Rate). We use the same attack settings as used in prior work~\cite{ma2019nic,xu2017feature}.}\label{tab:comp}
\begin{tabular}{c||c|c|c|c||c|c|c|c}
\specialrule{.1em}{.05em}{.05em}
\textbf{Dataset}  & \multicolumn{4}{c||}{CIFAR-10} & \multicolumn{4}{c}{ImageNet}   \\ \hline
\textbf{Detector} & \oursys & NIC   & FS    & LID   & \oursys & NIC    & FS    & LID    \\ 
\specialrule{.1em}{.05em}{.05em}
\textbf{FPR}      & \underline{0.6\%} & 4.2\% & 5.6\% & 4.9\% & \underline{2.7\%} & 14.6\% & 8.3\% & 14.5\% \\ \hline
\textbf{DR: CW-$L_2$}    & \underline{100\%} & 96\%  & 100\% & 86\%  & \underline{98.9\%}  & 96\%   & 92\%  & 78\%   \\ \hline
\textbf{DR: DFool}    & \underline{99.4\%}  & 91\%  & 77\% & 84\%  & \underline{95.0\%}  & 92\%   & 79\%  & 83\%   \\ 
\specialrule{.1em}{.05em}{.05em}
\end{tabular}
\end{table*}

We use $\mathcal{D}_I$-\texttt{Train}, $\mathcal{D}_C$-\texttt{Carlini}-\texttt{Train}, and $\mathcal{D}_C$-\texttt{ResNet}-\texttt{Train} (see Section~\ref{sec:setup}) to train our detectors and evaluate them based on the corresponding testing sets. 

\noindent \textbf{{CW-$L_2$ attacks.}}
As shown in Table~\ref{tab:cw}, the proposed technique achieves very high detection rates (up to 100\% on CIFAR-10, and 99.3\% on ImageNet) 
with low FPR values. The results are stable across different target models, confidence levels, and classification methods. 

In addition to SVM and Adaboost, we also train a fully connected neural network as the AE classifier, and obtain very similar results.  It shows that it does not affect the performance by using a more sophisticated classifier. It also indicates that the effect of E\&R does not depend on a specific classifier type.

\vspace{3pt}
\noindent \textbf{{DeepFool attacks.}}
For another leading $L_2$ AE generation algorithm---DeepFool (see Section~\ref{sec:deepfool}), we observe very similar results as CW-$L_2$. Table~\ref{tab:cw} shows that our detector achieves very high detection rates (up to 99.8\% on CIFAR-10, and 95.0\% on ImageNet) with low FPR values. 

%
\begin{figure}
\subfloat[SVM]{\includegraphics[scale=0.25,trim=10 0 0 0, clip]{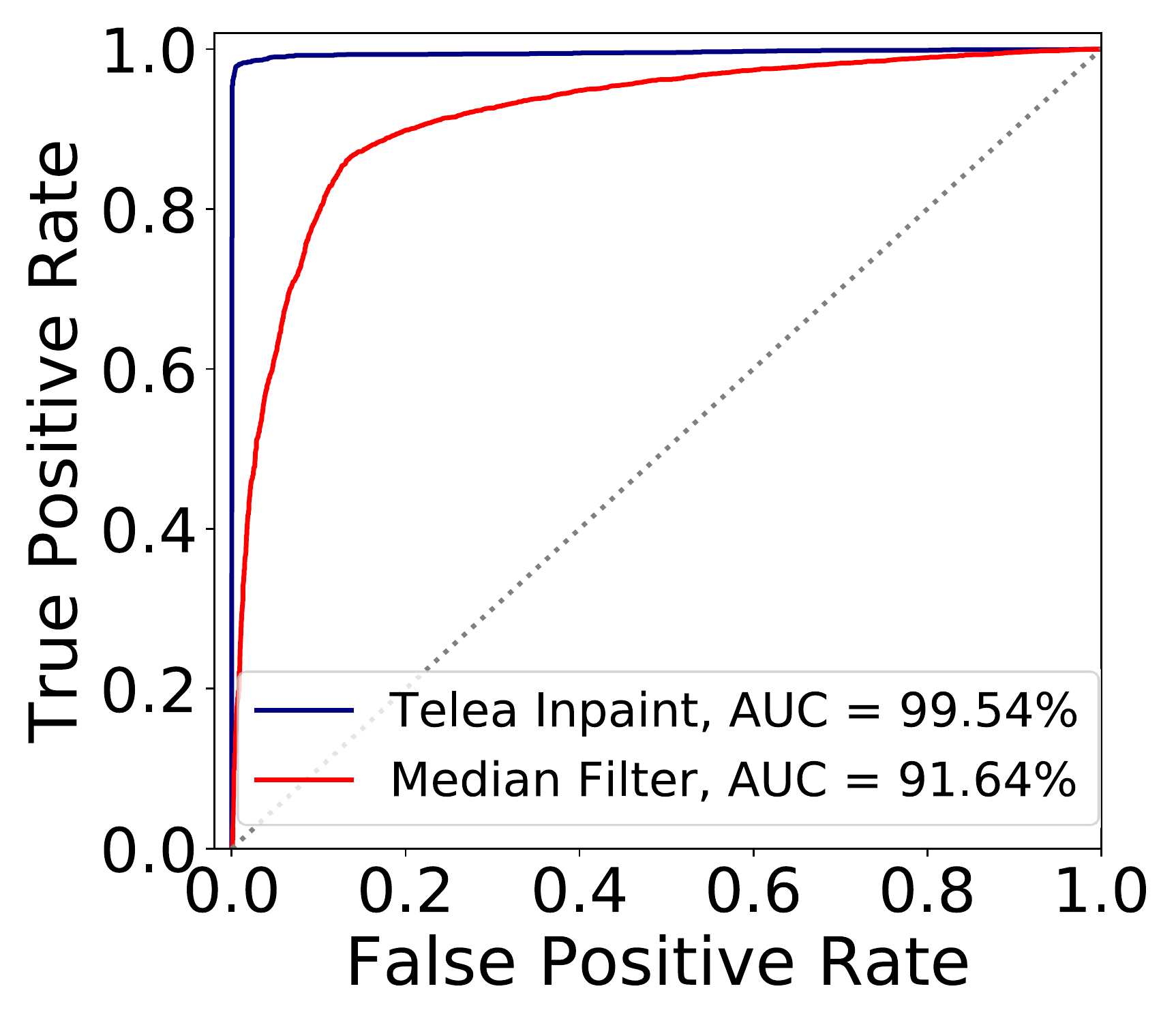}}
\subfloat[AdaBoost]{\includegraphics[scale=0.25,trim=10 0 0 0, clip]{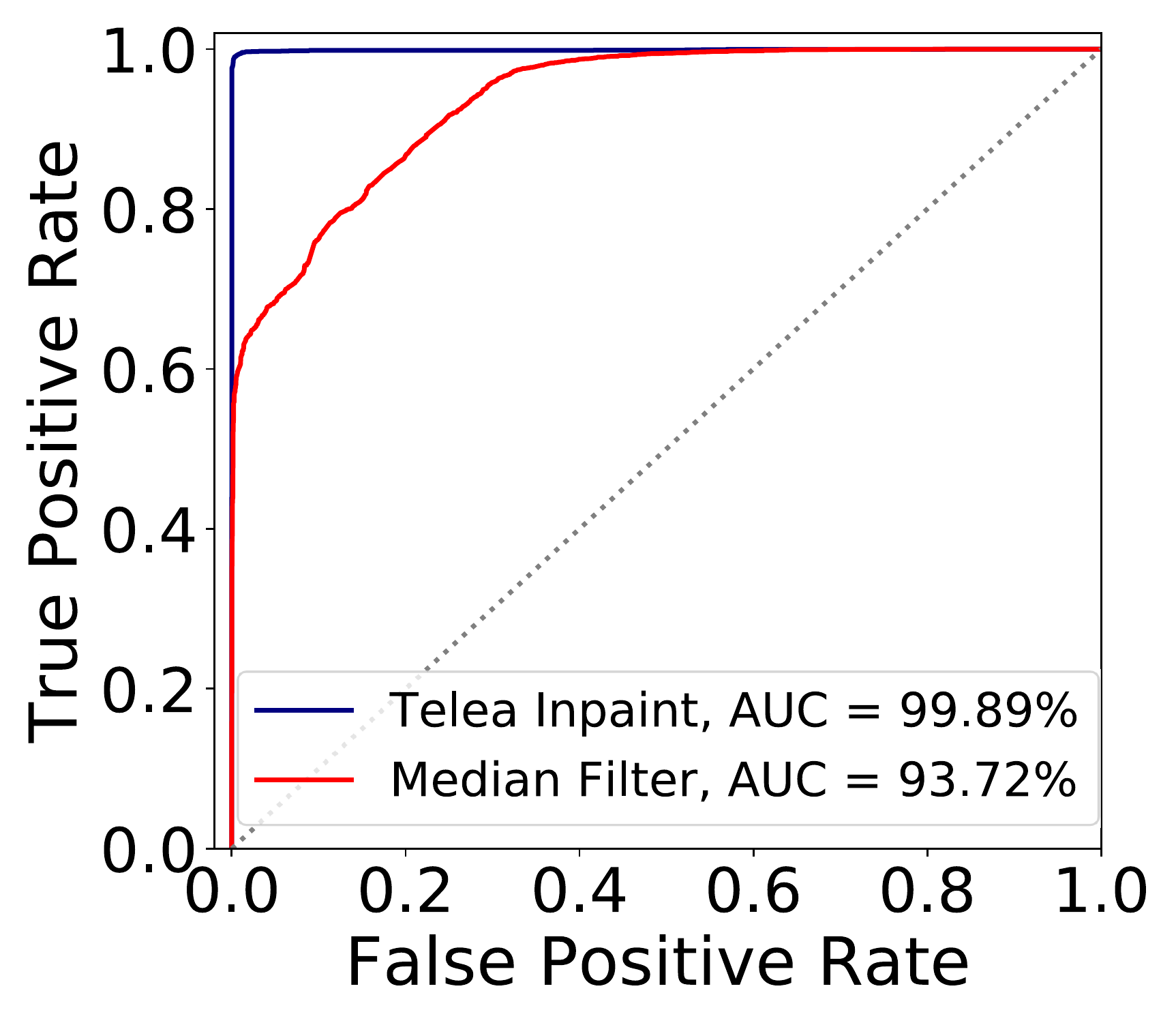}}     
\caption{ROC curves.}
\label{fig:roc_detector}
\end{figure}

\vspace{3pt}
\noindent \textbf{{Comparison with baseline.}} 
To illustrate the the benefits of the Telea inpainting algorithm used in our detector, 
we compare it with a baseline method, which
uses a median filter to recover the damaged pixels. 
In particular, the window size of our median filter is 3$\times$3, which is also adopted by Feature Squeezing~\cite{xu2017feature}. Without loss of of generality, the datasets we use are $\mathcal{D}_C$-\texttt{ResNet}-\texttt{Train} and $\mathcal{D}_C$-\texttt{ResNet}-\texttt{Test}. 
We replace the Telea inpainting with the median filter in our implementation to build a
baseline detector.
Figure~\ref{fig:roc_detector} shows the comparison result using ROC (receiver operating characteristic) curves of the different detectors. As shown in Figure~\ref{fig:roc_detector}(a), when SVM is used as the classifier, the AUC value declines from 99.54\% to 91.64\%. Similarly, as shown in Figure~\ref{fig:roc_detector}(b), when AdaBoost is used, the AUC value correspondingly declines from 99.89\% to 93.72\%. Thus, a high-quality inpainting method is closely related to the final performance of our AE detector.

\vspace{3pt}
\noindent \textbf{{Comparison with prior work.}}
As summarized in Table~\ref{tab:comp},
we compare \oursys with some state-of-the-art AE detectors---NIC~\cite{ma2019nic}, LID~\cite{ma2018characterizing}, and Feature Squeezing~\cite{xu2017feature}. 
\fedit{For CW-$L_2$ attack, their experiments only examine $\kappa$ = 0.0, which is the default setting, so we also list the results under $\kappa$ = 0.0 in Table~\ref{tab:comp} (see Table~\ref{tab:cw} for the results of our detector under other $\kappa$ values). We take NIC as an example here. With respect to CIFAR-10, NIC obtains the detection rate 
 96\% (see Table I in~\cite{ma2019nic}), while our system achieves the detection rate \textbf{100\%}. With respect to ImageNet, the detection rate of NIC 
 is 96\% (see Table I in~\cite{ma2019nic}), while our detection rate is \textbf{98.9\%}. In terms of DeepFool, \oursys also outperforms other AE detectors. 
When considering CIFAR-10, our system obtains the detection rate \textbf{99.4\%}, while NIC~\cite{ma2019nic} obtains the detection rate 91.0\% (see Table I in~\cite{ma2019nic}). Similarly, when considering ImageNet, \oursys can achieve the detection rate \textbf{95.0\%}, that is superior to NIC, the detection rate of which is 92\%.}

\fedit{More importantly, from the angle of FPR, the performance of \oursys is significantly better than other detectors. For example, when considering CIFAR-10, the FPR of NIC is 4.2\%, while ours is \textbf{0.6\%}. Moreover, when considering ImageNet, the FPR of NIC is 14.6\%, while ours is only \textbf{2.7\%}. 
It is worth noting that the distribution of
adversarial and benign images is not balanced in practice---most
inputs should be benign. Thus, FPR is a very important metric to evaluate 
the model performance: a lower FPR indicates that the system makes fewer mistakes 
for benign images. \oursys is able to keep both a high detection rate and a 
\emph{very low FPR}. }

\subsection{Notable Characteristics}
\vspace{3pt}
\noindent \textbf{{Target-model agnostic.}}
We are interested in finding out whether a detector trained using AEs targeting one model 
can be directly used to detect AEs targeting another---that is, whether it is 
\emph{target-model agnostic}. We thus train our system using CW-$L_2$ AEs in
$\mathcal{D}_C$-\texttt{Carlini}-\texttt{Train}, 
and test it using CW-$L_2$ AEs in $\mathcal{D}_C$-\texttt{ResNet}-\texttt{Test}. 

\begin{table}[!th]
\centering
\caption{Target-model agnostic property of \oursys.}\label{tab:cw_2}

\begin{tabular}{c|c||c|c|c}
\specialrule{.1em}{.05em}{.05em}
      \textbf{Target Model} & \multirow{2}{*}{\textbf{Classifier}} &
      \multicolumn{3}{c}{\textbf{Detection Rate}}  \\ \cline{3-5}
      
  \textbf{(Train $\rightarrow$ Test)} & & $\kappa$=0.0 & $\kappa$=0.4 & $\kappa$=1.0 \\ \specialrule{.1em}{.05em}{.05em}
  \multirow{2}{*}{Carlini$\rightarrow$ResNet32}  & {SVM} & {100\%} & {100\%} & {100\%} \\ \cline{2-5} 
& AdaBoost & 97.9\%   & 97.9\% & 98.2\% \\ \hline
  \multirow{2}{*}{ResNet32$\rightarrow$Carlini}  & {SVM}   &  {99.9\%}   & {99.9\%} & {99.8\%} \\ \cline{2-5}
                      & AdaBoost    & 99.7\%   & 99.8\% & 99.6\% \\
\specialrule{.1em}{.05em}{.05em}
\end{tabular}
\end{table}

As Table~\ref{tab:cw_2} shows, the detection rate is as high as 100\%. We then 
train the system using CW-$L_2$ AEs in
$\mathcal{D}_C$-\texttt{ResNet}-\texttt{Train}, and test it using CW-$L_2$ AEs in $\mathcal{D}_C$-\texttt{Carlini}-\texttt{Test}; 
the detection rate is as high as 99.9\%.

Therefore, this experiment not only confirms that \oursys is \textit{target-model 
agnostic}, but also demonstrates that \oursys has low risk of overfitting.

\vspace{3pt}
\noindent \textbf{{Transferability.}}
We are also interested in the transferability of our detector---whether \oursys trained on one type of
AEs can be directly applied to detect another type of
AEs that are \emph{unseen} during training. To verify it, we \emph{train} \oursys using CW-$L_2$ AEs in 
$\mathcal{D}_C$-\texttt{Carlini}-\texttt{Train}, without loss of generality. 
Then, we test the trained system using DeepFool AEs in $\mathcal{D}_C$-\texttt{ResNet}-\texttt{Test} and 
$\mathcal{D}_C$-\texttt{Carlini}-\texttt{Test} , and our system can achieve detection rates 97.1\% and 96.2\%, respectively. 
Thus, we can conclude the proposed technique has very good transferability, that is,
it keeps effective in handling unseen AE generation methods.

\begin{figure*}[t]

\centering
\includegraphics[scale=0.5]{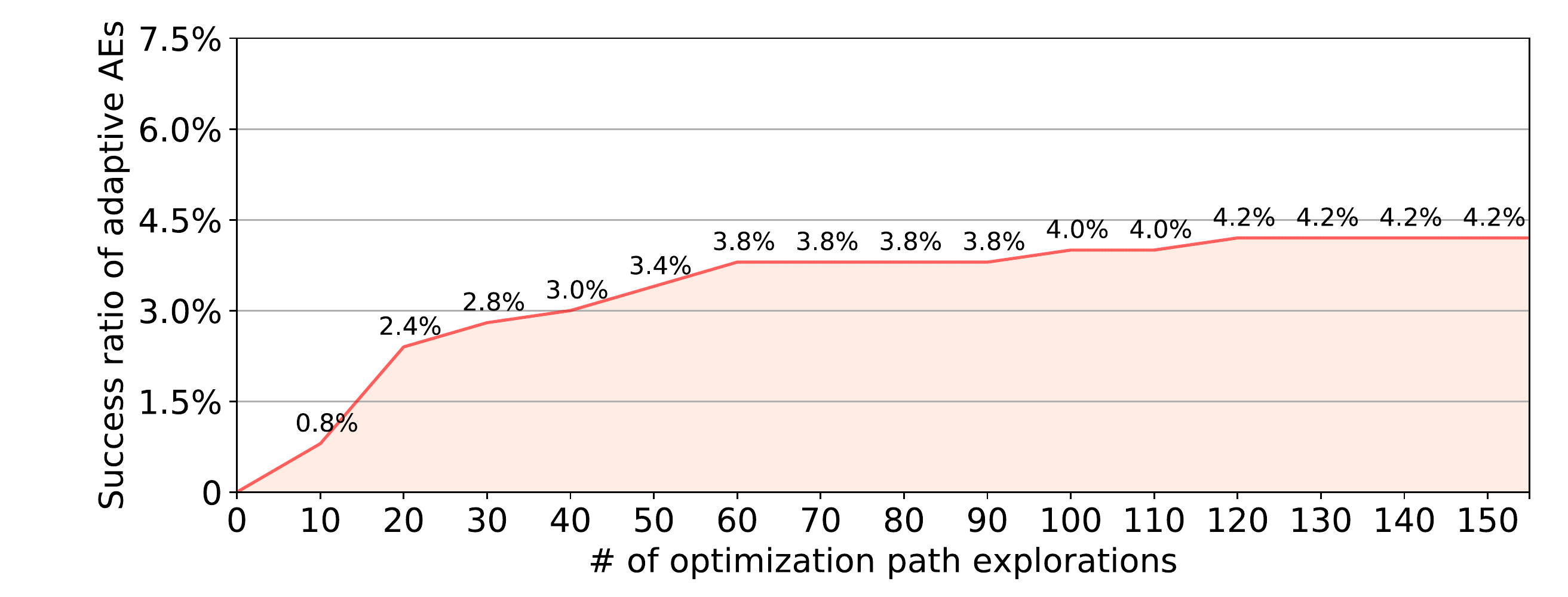}

\caption{Success ratio of adaptive AEs.}\label{fig_adapt}

\end{figure*}

\vspace{3pt}
\noindent  \textbf{{Explanation.}} The two notable properties of \oursys---target-model agnostic and
good transferability---can be attributed to the unique advantage of the proposed approach:
benign samples and 
AEs show distinct sensitivities to the E\&R operations, which
do not depend on the target model and the attack method. 

\subsection{Value Selection for the Parameter $n$.}\label{sec:n}


\begin{table}[!thb]\small
\centering
\caption{Impacts of different values of $n$ (CIFAR-10).}\label{tab:par_cifar}
\begin{tabular}{c|c|c|c|c|c|c}
\specialrule{.1em}{.05em}{.05em}  
 \multicolumn{1}{c|}{\multirow{2}{*}{\begin{tabular}[c]{@{}c@{}}\textbf{Target}\\ \textbf{Model}\end{tabular}}} & \multirow{2}{*}{\textbf{Classifier}}   & \multicolumn{1}{c|}{\multirow{2}{*}{\textbf{FPR}}} & \multicolumn{3}{c|}{\textbf{Detection Rate}}                                                     & \multicolumn{1}{c}{\multirow{2}{*}{}}    \\ \cline{4-6}
 \multicolumn{1}{c|}{}                                                                        &                               & \multicolumn{1}{c|}{}                     & \multicolumn{1}{c|}{$\kappa$=0.0}  & \multicolumn{1}{c|}{$\kappa$=0.4}  & \multicolumn{1}{c|}{$\kappa$=1.0}  & \multicolumn{1}{c}{}             \\ \specialrule{.1em}{.05em}{.05em}  

\multicolumn{1}{c|}{\multirow{2}{*}{Carlini}}                                                & SVM                           & \multicolumn{1}{c|}{0.4\%}                & \multicolumn{1}{c|}{100\%}  & \multicolumn{1}{c|}{100\%}  & \multicolumn{1}{c|}{100\%}  & \multicolumn{1}{c}{\multirow{4}{*}{\rotatebox{45}{n=3}}} \\ \cline{2-6}

\multicolumn{1}{c|}{}                                                                        & \multicolumn{1}{l|}{Adaboost} & 0.0\%                                     & 100\%                       & 100\%                       & 99.9\%                      & \multicolumn{1}{c}{}                     \\ \cline{1-6}

\multicolumn{1}{c|}{\multirow{2}{*}{ResNet32}}                                               & SVM                           & \multicolumn{1}{c|}{3.6\%}                & \multicolumn{1}{c|}{99.6\%} & \multicolumn{1}{c|}{99.6\%} & \multicolumn{1}{c|}{99.6\%} & \multicolumn{1}{c}{}                     \\ \cline{2-6}

 \multicolumn{1}{c|}{}                                                                        & \multicolumn{1}{l|}{Adaboost} & 0.9\%                                     & 99.2\%                      & 99.1\%                      & 98.5\%                      & \multicolumn{1}{c}{}                     \\
\cline{1-7}  

\multirow{2}{*}{Carlini}                                                                     & SVM                           & 0.4\%                                     & 100\%                       & 100\%                       & 100\%                       & \multirow{4}{*}{\rotatebox{45}{n=5}}                      \\ \cline{2-6}
                                         & Adaboost                      & 0.0\%                                     & 100\%                       & 99.9\%                      & 99.9\%                      &                                           \\ \cline{1-6}
 \multirow{2}{*}{ResNet32}                                                                    & SVM                           & 3.3\%                                     & 99.6\%                      & 99.6\%                      & 99.6\%                      &                                           \\ \cline{2-6}
                                             & Adaboost                      & 0.7\%                                     & 99.2\%                      & 99.1\%                      & 99.0\%                      &                                           \\ \cline{1-7}

\multirow{2}{*}{Carlini}                                                                     & SVM                           & 0.4\%                                     & 100\%                       & 100\%                       & 100\%                       & \multirow{4}{*}{\rotatebox{45}{n=7}}                      \\ \cline{2-6}
                                         & Adaboost                      & 0.0\%                                     & 100\%                       & 100\%                      & 99.8\%                      &                                           \\ \cline{1-6}
 \multirow{2}{*}{ResNet32}                                                                    & SVM                           & 2.9\%                                     & 99.6\%                      & 99.7\%                      & 99.7\%                      &                                           \\ \cline{2-6}
                                             & Adaboost                      & 0.9\%                                     & 99.3\%                      & 99.1\%                      & 98.9\%                      &                                           \\ \cline{1-7}

\multirow{2}{*}{Carlini}                                                                     & SVM                           & 0.4\%                                     & 100\%                       & 100\%                       & 100\%                       & \multirow{4}{*}{\rotatebox{45}{n=9}}                      \\ \cline{2-6}
                                         & Adaboost                      & 0.0\%                                     & 100\%                       & 100\%                      & 99.8\%                      &                                           \\ \cline{1-6}
 \multirow{2}{*}{ResNet32}                                                                    & SVM                           & 3.0\%                                     & 99.7\%                      & 99.7\%                      & 99.7\%                      &                                           \\ \cline{2-6}
                                             & Adaboost                      & 0.7\%                                     & 99.3\%                      & 99.3\%                      & 99.1\%                      &                                           \\ \cline{1-7}

 \multirow{2}{*}{Carlini}                                                                     & \multicolumn{1}{c|}{SVM}      & 0.4\%                                     & 100\%                       & 100\%                       & 100\%                       & \multirow{4}{*}{\rotatebox{45}{n=11}}                     \\ \cline{2-6}
                                                          & \multicolumn{1}{c|}{Adaboost} & 0.0\%                                     & 99.8\%                       & 99.9\%                       & 99.7\%                      &                                           \\ \cline{1-6}
                                                 \multirow{2}{*}{ResNet32}                                                                    & \multicolumn{1}{c|}{SVM}      & 2.8\%                                     & 99.6\%                      & 99.7\%                      & 99.7\%                      &                                           \\ \cline{2-6}
                                                                                                                     & \multicolumn{1}{c|}{Adaboost} & 0.8\%                                     & 99.0\%                      & 99.2\%                      & 98.9\%                      &                                      \\  \specialrule{.1em}{.05em}{.05em}  
\end{tabular}
\end{table}

\begin{table}[!thb]\small
\centering
\caption{Impacts of different values of $n$ (ImageNet).}\label{tab:par_imgNet}
\begin{tabular}{c|c|c|c|c|c|c}
\specialrule{.1em}{.05em}{.05em}  

 \multicolumn{1}{c|}{\multirow{2}{*}{\begin{tabular}[c]{@{}c@{}}\textbf{Target}\\ \textbf{Model}\end{tabular}}} & \multirow{2}{*}{\textbf{Classifier}}   & \multicolumn{1}{c|}{\multirow{2}{*}{\textbf{FPR}}} & \multicolumn{3}{c|}{\textbf{Detection Rate}}                                                     & \multicolumn{1}{c}{\multirow{2}{*}{}}    \\ \cline{4-6}
 \multicolumn{1}{c|}{}                                                                        &                               & \multicolumn{1}{c|}{}                     & \multicolumn{1}{c|}{$\kappa$=0.0}  & \multicolumn{1}{c|}{$\kappa$=0.4}  & \multicolumn{1}{c|}{$\kappa$=1.0}  & \multicolumn{1}{c}{}                     \\ \specialrule{.1em}{.05em}{.05em}  

 \multicolumn{1}{c|}{\multirow{10}{*}{ResNet50}}                                               & SVM                           & \multicolumn{1}{c|}{9.8\%}                & \multicolumn{1}{c|}{95.4\%} & \multicolumn{1}{c|}{95.1\%} & \multicolumn{1}{c|}{95.5\%} & \multirow{2}{*}{\rotatebox{45}{n=3}}                      \\ \cline{2-6}
 \multicolumn{1}{c|}{}                                                                        & \multicolumn{1}{l|}{Adaboost} & 6.6\%                                     & 93.1\%                      & 91.4\%                      & 93.8\%                      &                     \\ \cline{2-7}

                                                                   & SVM                           & 4.7\%                                     & 95.5\%                      & 95.8\%                      & 97.3\%                      &    \multirow{2}{*}{\rotatebox{45}{n=5}}                                        \\ \cline{2-6}
 & Adaboost                      & 2.8\%                                     & 96.5\%                      & 97.6\%                      & 97.2\%                      &                                           \\  \cline{2-7}
                                                                   & \multicolumn{1}{c|}{SVM}      & 3.6\%                                     & 97.6\%                      & 98.1\%                      & 98.2\%                      &       \multirow{2}{*}{\rotatebox{45}{n=7}}                                     \\ \cline{2-6}
           & \multicolumn{1}{c|}{Adaboost} & 2.1\%                                     & 97.9\%                      & 98.6\%                      & 98.6\%                      &                                           \\ \cline{2-7}
                                                                   & \multicolumn{1}{c|}{SVM}      &  3.5\%                                     &  97.6\%                      &  98.0\%                      &  98.3\%                      &       \multirow{2}{*}{\rotatebox{45}{n=9}}                                     \\ \cline{2-6}
           & \multicolumn{1}{c|}{Adaboost} &  2.0\%                                     & 98.0\%                      &  98.4\%                      &  98.8\%                      &                                           \\ \cline{2-7}
                                                                  & \multicolumn{1}{c|}{SVM}      & 3.2\%                                     & 97.6\%                      & 98.1\%                      & 98.5\%                      &       \multirow{2}{*}{\rotatebox{45}{n=11}}                                     \\ \cline{2-6}
           & \multicolumn{1}{c|}{Adaboost} & 1.4\%                                     & 98.4\%                      & 98.5\%                      & 98.9\%                      &                                           \\

           \specialrule{.1em}{.05em}{.05em}  
\end{tabular}
\end{table}

We use $n=11$ in the previous experiments.
Here, we investigate the
impacts of different values of $n$ on the detector's performance. 
The CW-$L_2$ AEs in   $\mathcal{D}_C$-\texttt{Carlini}-\texttt{Train}, $\mathcal{D}_C$-\texttt{ResNet}-\texttt{Train}, and $\mathcal{D}_I$-\texttt{Train} are used in this experiment. 
For CIFAR-10, which has only 10 classes (thus no PCA is needed), 
varying the value of $n$ has little impacts.
However, for ImageNet, the value of $n$ has 
noticeable impacts: when $n$ increases, the AE detection rate
increases and FPR decreases (see Table~\ref{tab:par_cifar} and Table~\ref{tab:par_imgNet} for more details). The reason is that by increasing $n$,
more principal components can be extracted (see Section 4).
However, when $n>11$, the performance improvement is negligible, probably because the extra principal components do not provide useful features for AE detection.
Therefore, we adopt $n=11$. 

\subsection{Efficiency of \oursys}
We investigate the efficiency of the proposed technique on ImageNet because large-sized images consume more processing time. For a single image, ResNet50 needs approximately 1.076 seconds for classification. Since parallel computing is supported by GPU, given a relatively small number of images as inputs (e.g., $n=11$), it takes similar time to generate the classification vectors for them. Apart from this, 
to detect AE, our method brings additional 1.01
seconds by average. In detail, it consumes 0.264 seconds for the inpainting, 0.744 seconds for the PCA-based dimension reduction, and 0.002 seconds for the final prediction (taking SVM as an example). In short,  our detector causes a small delay.

\section{Resilience to Adaptive Attacks}\label{sec:adapt}
In an adaptive attack threat model, an adversary knows the existence and internal 
details of our detector and \emph{adapts} the attacks to bypass the detection. We thus seek to 
study the resilience of \oursys to adaptive attacks. 

An AE detector can be categorized as either differentiable or non-differentiable. 
Several previous works propose defense mechanisms that apply differentiable transformations to an image before detection or classification~\cite{metzen2017detecting,gong2017adversarial,grosse2017statistical,tian2018detecting}. But attackers can circumvent 
these  differentiable defenses by ``\emph{differentiating through them}''---\emph{i.e.,} by taking the gradient
of a class probability regarding input pixels through both the CNN and the transformation~\cite{prakash2018deflecting,carlini2017adversarial,he2017adversarial}. This strategy, however, is \emph{inapplicable} to bypassing \oursys. 
Due to the random-erasing and inpainting-based restoring, our approach is not only non-differentiable but involves tremendous randomness. 

To bypass non-differentiable defences, Backward Pass Differentiable Approximation (BPDA) is proposed~\cite{obfuscated-gradients}. To handle defenses that employ 
randomized transformation to the input (like ours), it applies 
Expectation over Transformation~\cite{athalye2018synthesizing} to compute the gradient over the expected
transformation to the input. However, in our approach the erased pixels 
are randomly selected among all the image pixels, and there are 
around 7.7$\times 10^{140}$ unique masks  (even for a small image; see Section~\ref{sec:details}); 
thus, it is infeasible to
calculate the expected transformation. Moreover, \oursys is not only
randomized but also non-differentiable; in this case, it is unknown
how to apply BPDA to bypassing \oursys.

\vspace{3pt}
\noindent \textbf{Adaptive AE generation.}
He et al.~\cite{he2017adversarial} describe a representative adaptive attack method
against non-differentiable defences, where 
an attacker tries to circumvent the defensive approach by (a) considering intermediate distorted images during optimization
and (b) exploring multiple diverse 
optimization paths. 

Inspired by \cite{he2017adversarial},  we design similar adaptive attacks to examine the resilience of 
our approach.
To that end, we modify the code of the CW algorithm~\cite{carlini2017towards}, in order to adaptively generate AEs that can bypass our detector. 
Specifically, after each iteration in an optimization procedure, an intermediate 
distorted image is obtained. 
We then check 
whether it can bypass our detector. For each image,
we repeat the optimization procedure for up to $T$ times to explore different optimization paths 
(for this purpose, we set a randomly initialized state at the beginning of each optimization procedure). 
As shown in Figure~\ref{fig_adapt}, we set $T=150$, corresponding to
around 450 seconds on average on our machine.
In comparison,
the two works~\cite{tian2018detecting} and~\cite{he2017adversarial} use around 
75 and 180 seconds to generate adaptive AEs for each image, respectively.  

Given that adaptive CW AE generation is quite time-consuming, without loss of generality,
this experiment is conducted on 500 images randomly selected from CIFAR-10.
During the AE generation, we 
let $\kappa=0.0$, which means that the resulting AE is classified as the target class. 
As $\kappa$ increases, the model classifies the resulting AE as the attacker-desired label more likely. 
As a larger value of $\kappa$ imposes an extra constraint to attackers and lowers the
chance of successful adaptive attacks, we only consider $\kappa=0.0$.

\vspace{3pt}
\noindent \textbf{Resilience results.}
We adopt the SVM-based detector that achieves a detection rate of 100\%  (Table~\ref{tab:cw}): no AEs can fool it \emph{without adaptive attacks}. 
Figure shows that only 4.2\% (that is, 21 AEs) of adaptive AEs can bypass our detector. By contrast, similar adaptive attacks~\cite{he2017adversarial} 
can bypass Feature Squeezing based AE detection~\cite{xu2017feature} at a success rate of 100\%;
as another example, \cite{tian2018detecting} can merely achieve a detection rate of 70\% under adaptive CW attacks.
More importantly,  the first 50 times of the optimization path exploration attain the success
rate of 3.4\%, while the following 100 times only increase the success rate by 0.8\%.
It shows that the effect of adaptive attacks grows very slowly as the attacker doubles his time. 
We thus can conclude that our detection technique is not only resilient to adaptive attacks 
based on differentiation, but also to adaptive attacks through exploration of many optimization paths. 
\zedit{Thus, \oursys, highly resilient to adaptive CW-$L_2$ attacks, fills a critical gap in AE detection.}

\section{Interpretability}\label{sec:interp}

\vspace{3pt}
\noindent\textbf{Background.} To make the final prediction, most neural-network-based image classifiers implement a \textit{softmax} function at the last layer
\begin{equation}
\begin{aligned}
&softmax(\mathbf{z})_i =\frac{e^{z_i}}{\sum^K_{j=1}e^{z_j}},\\ 
&\mathrm{for}\ i=1,\cdots,K\ \mathrm{and}\ \mathbf{z} =(z_1,\cdots,z_K)\in \mathbb{R}^K
\end{aligned}
\end{equation}
which maps an input vector $\mathbf{z}$ consisting $K$ real numbers
to a probability mass function over predicted output classes. The input vector of a \textit{softmax} function is also called \textit{logit}. Given a benign image whose logit is $\mathbf{z}$, the 
goal of an attacker is to perturb the image to get a new logit $\mathbf{z}'$ such that $\mathrm{argmax}_i(\mathbf{z}')\neq \mathrm{argmax}_i(\mathbf{z})$.

\vspace{3pt}
\noindent\textbf{Interpretation Using Classification Results.} Let $f(x)$ be the output of the \textit{softmax} layer of a neural network $f$ when feeding the input $x$. Let $T(x)$ be the output of processing $x$ with E\&R operations. If $x$ is benign, since it is not sensitive to E\&R operations, the probability mass functions $f(x)$ and $f(T(x))$ are similar. 
By contrast, if $x$ is an AE, $f(x)$ is significantly different from $f(T(x))$, since AEs are very sensitive to E\&R operations. In short, \zedit{if the sensitivity distinction between AEs and benign samples is true}, the divergence (or distance) between $f(x)$ and $f(T(x))$ should reflect 
whether $x$ is malicious or benign. \zedit{We then} \fedit{adopt} \zedit{two widely used metrics, Wasserstein distance (WD for short)~\cite{villani2009wasserstein}  and Kullback-Leibler divergence (KL for short)~\cite{kullback1997information}.}

\begin{figure}[!t]

\centering
\includegraphics[scale=0.4]{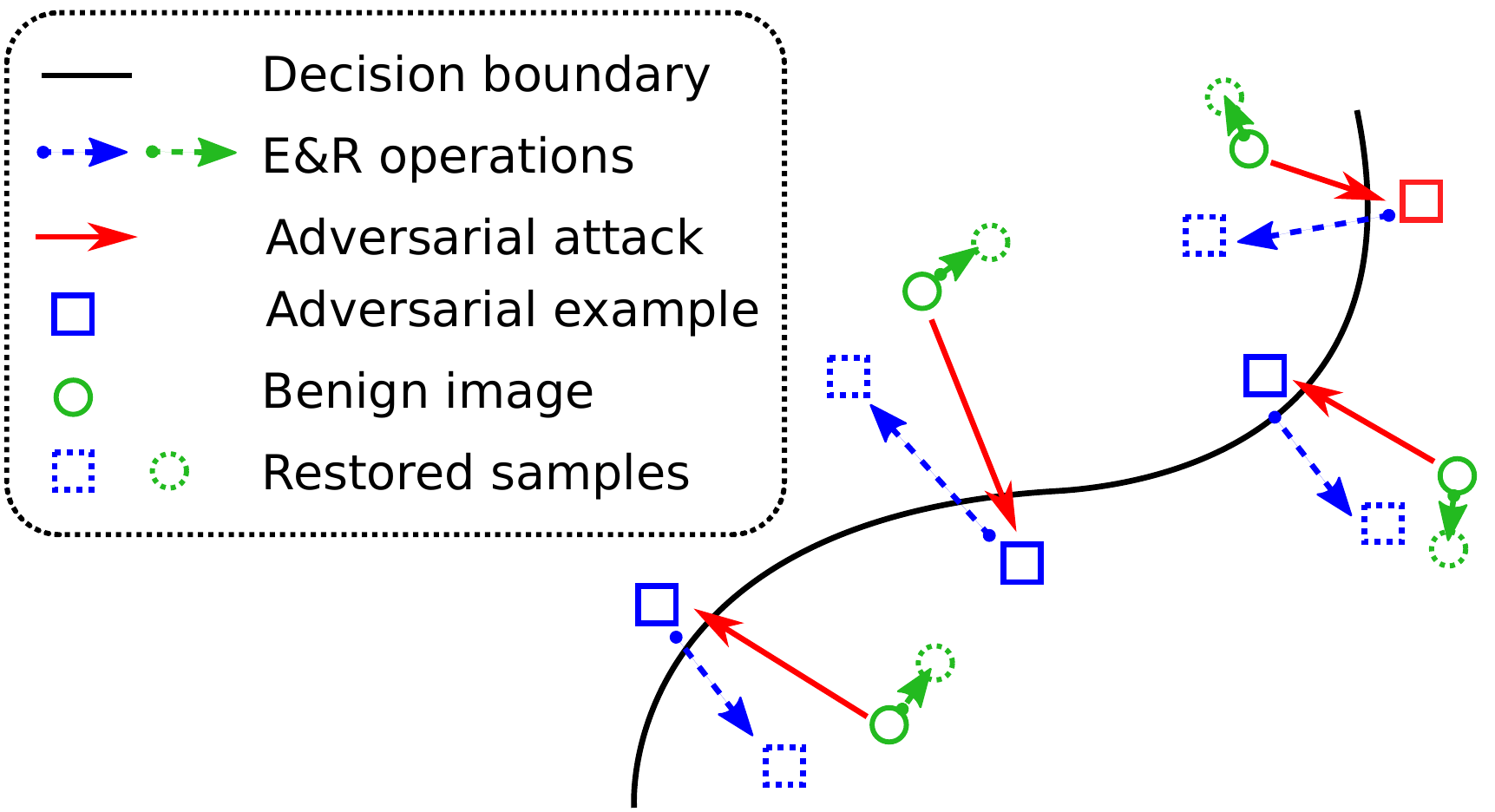}
\captionof{figure}{Illustration of how E\&R works.}\label{fig:interpret}

\end{figure}

\begin{table}
\caption{Clusters splitting result.}\label{tab:fpr_tpr}
\centering
\begin{tabular}{c|c|c|c}
\specialrule{.1em}{.05em}{.05em}
\multicolumn{1}{l|}{\textbf{Attacks}} & \textbf{Metrics} & \textbf{FPR}   & \multicolumn{1}{l}{\textbf{TPR}} \\ \specialrule{.1em}{.05em}{.05em}
\multirow{2}{*}{CW-$L_2$}           & WD      & 0.5\% & 78.4\%                              \\ \cline{2-4} 
                              & KL      & 0.0\% & 96.1\% \\                             \specialrule{.1em}{.05em}{.05em}
\multirow{2}{*}{DeepFool}        & WD      & 1.1\% & 85.7\%                              \\ \cline{2-4} 
                              & KL      & 0.5\% & 89.3\% \\                             \specialrule{.1em}{.05em}{.05em}
\end{tabular}

\end{table}


\begin{figure*}[!thb]
\centering
\subfloat[CW-$L_2$ (WD)]{\includegraphics[scale=0.3]{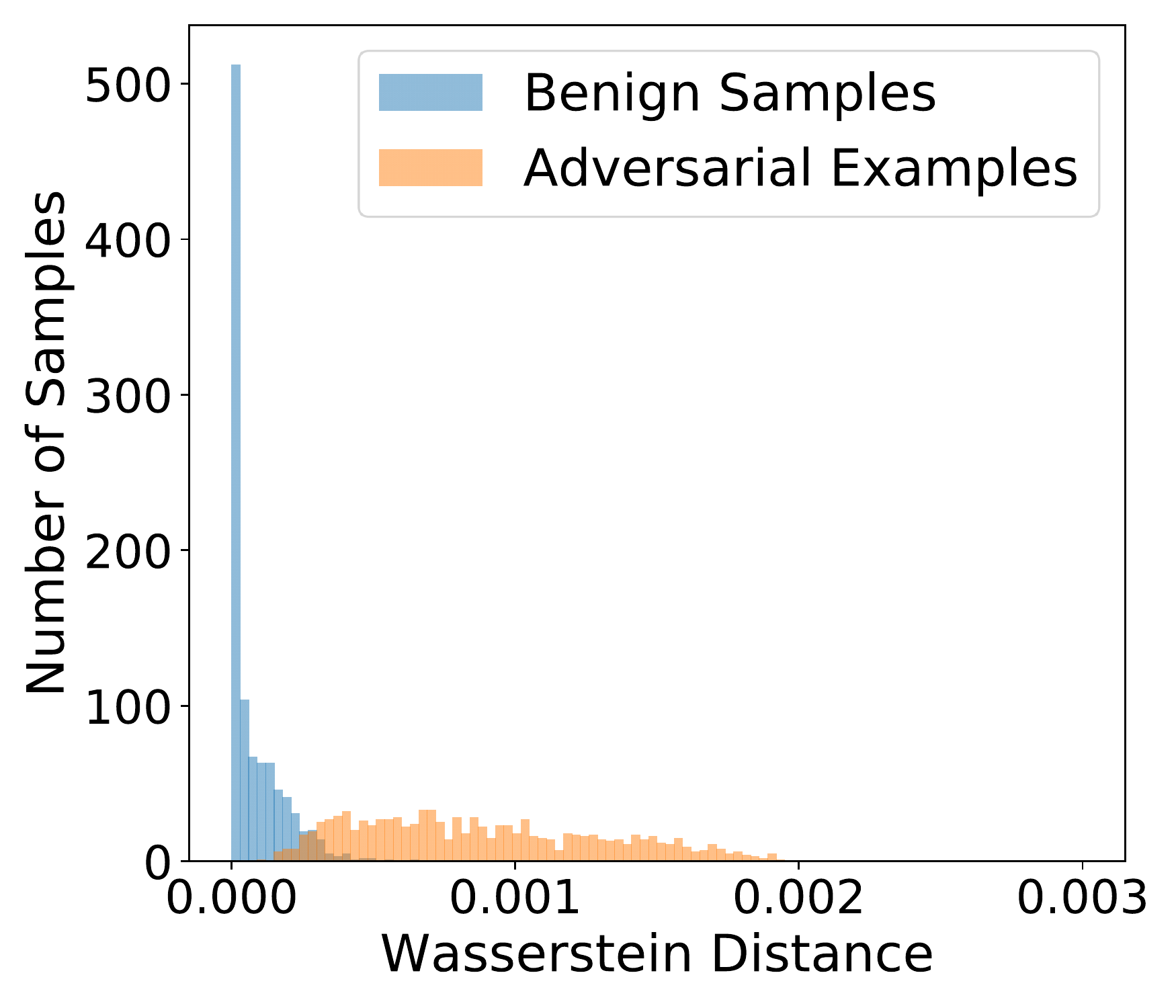}}\quad
\subfloat[CW-$L_2$ (KL)]{\includegraphics[scale=0.3]{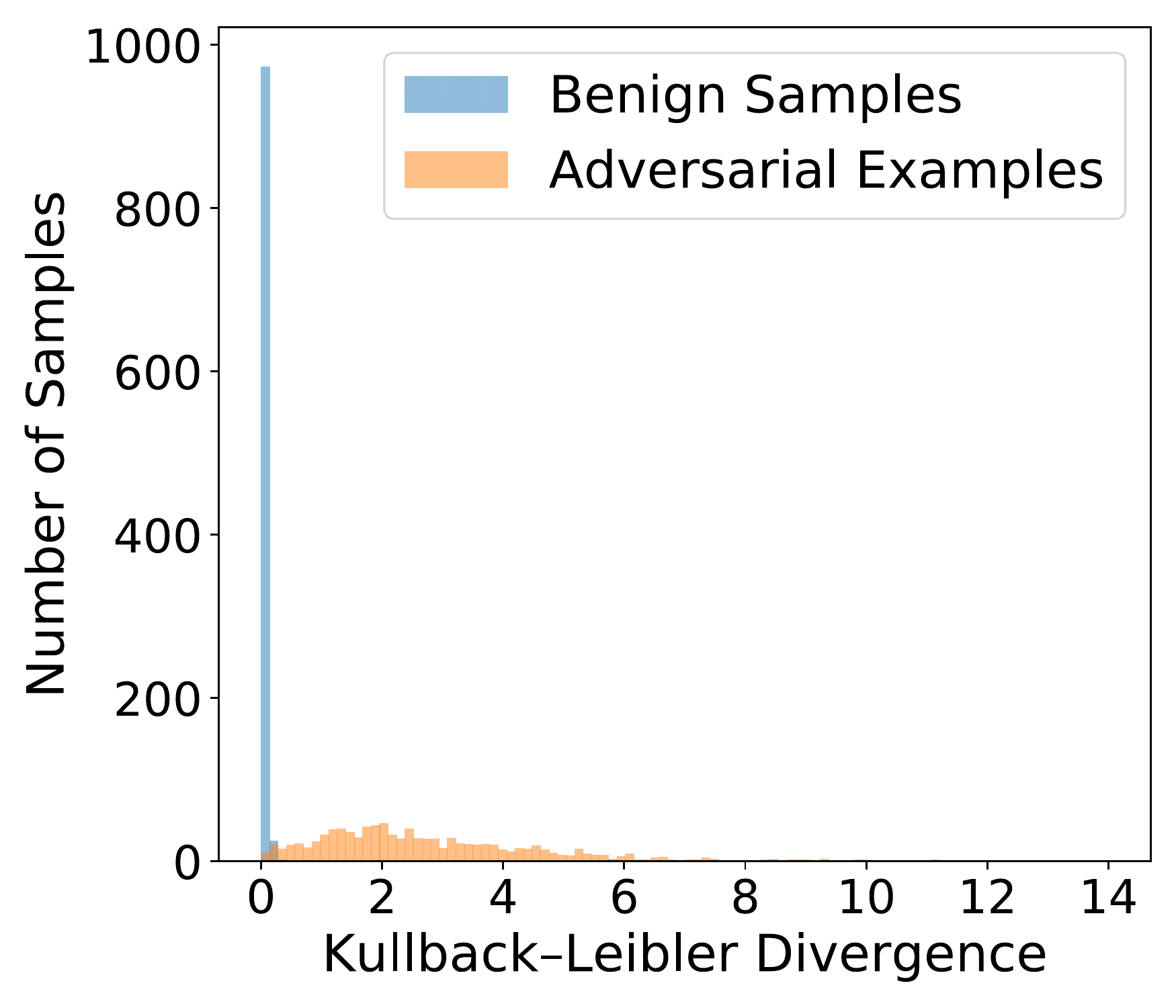}}\\
\subfloat[DeepFool (WD)]{\includegraphics[scale=0.3]{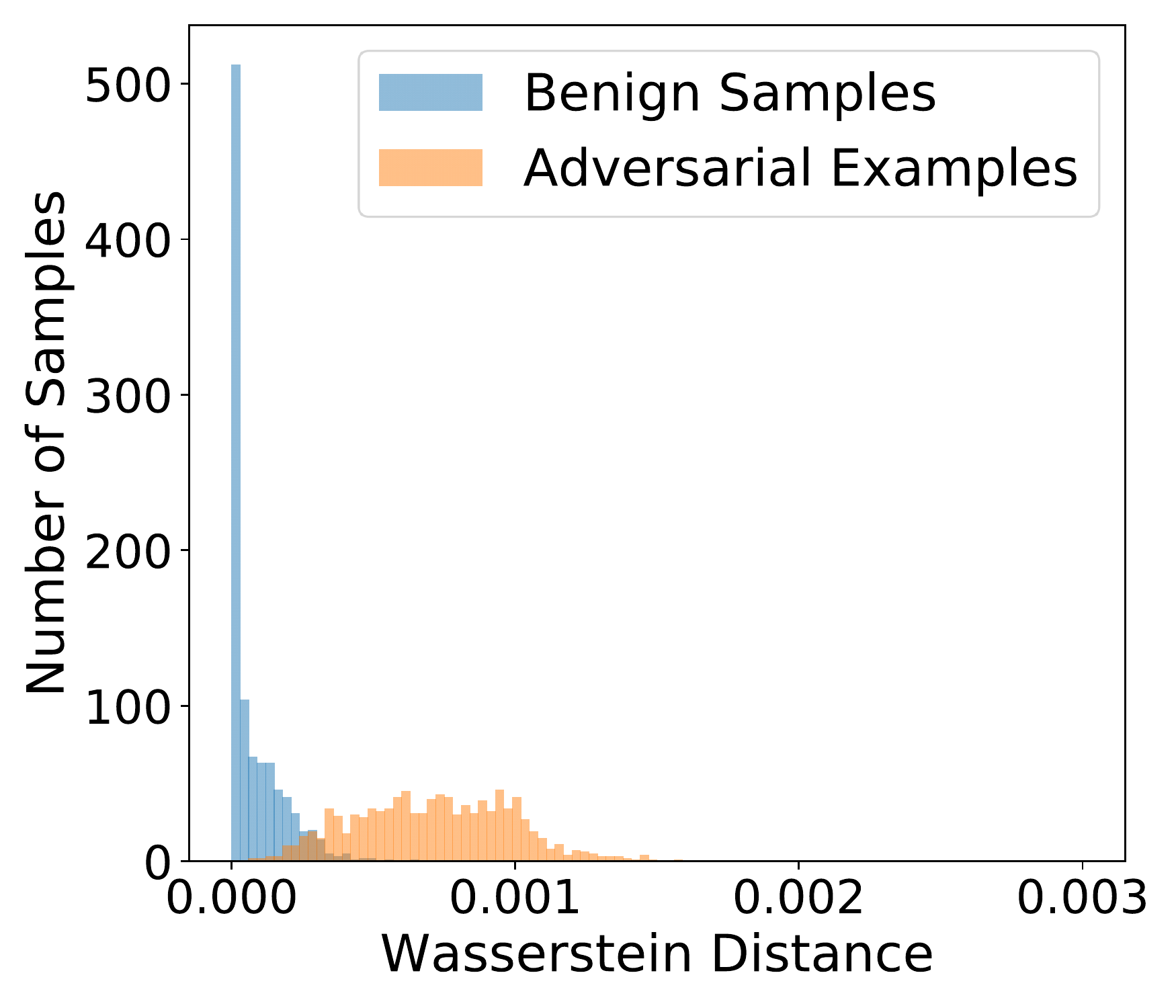}}\quad
\subfloat[DeepFool (KL)]{\includegraphics[scale=0.3]{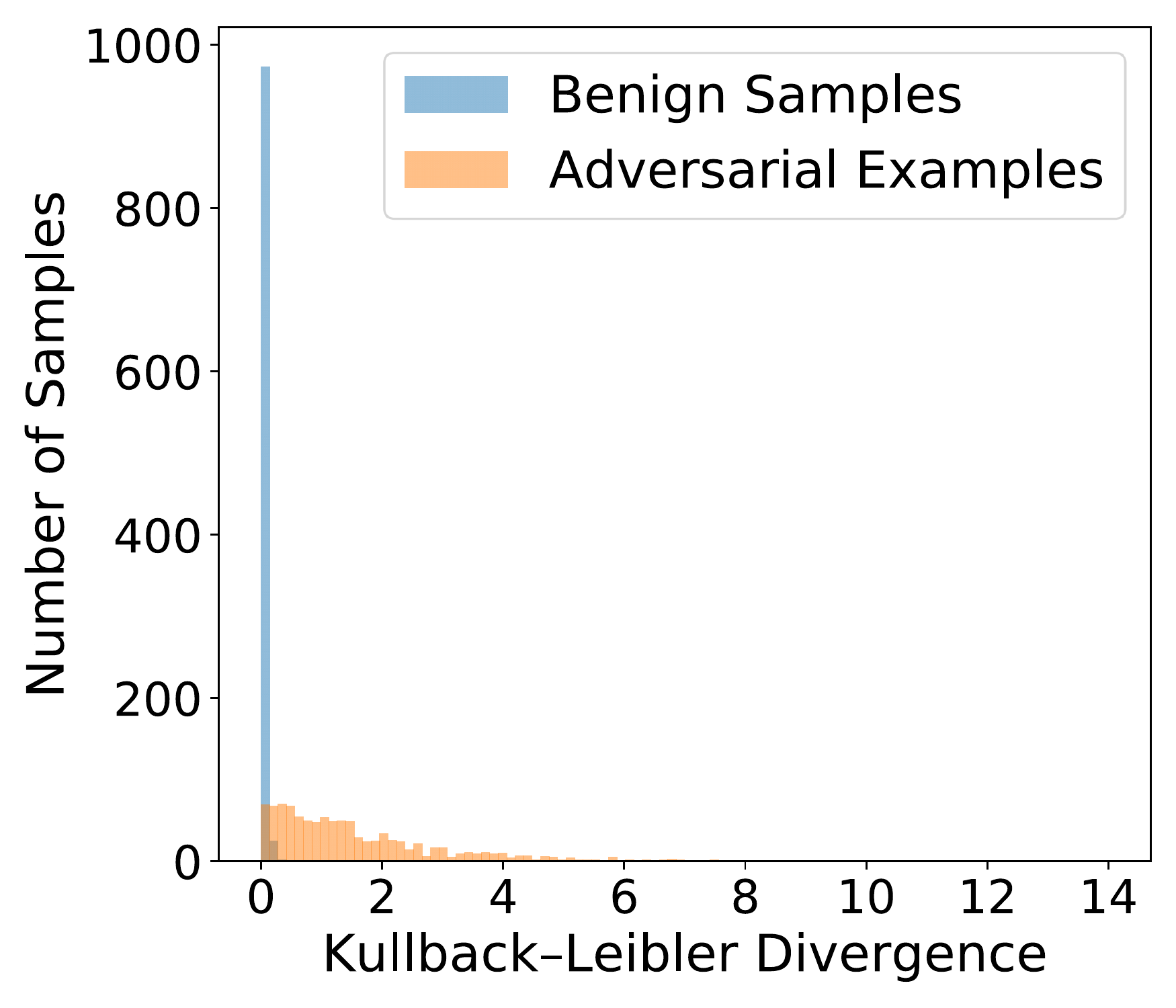}}

\caption{Visualization of the changes caused by E\&R on benign samples and AEs.}\label{wd_kl}
\end{figure*}

As shown in Figure~\ref{fig:interpret}, we depict benign and adversarial examples by green circles and blue squares, respectively. The arrows with dotted line represent E\&R operations. 
We consider the changes caused by E\&R operations on benign images and AEs (depicted by green and blue arrows with dotted line, respectively) should fall into different probability distributions. To visualize this, we randomly select 1,000 image pairs consisting of AEs and benign instances from $\mathcal{D}_I$-\texttt{Test}. After feeding them (with and without applying E\&R operations) into the image classification model, we collect the output of the \textit{softmax} layer. Then, we measure the difference between $f(x)$ and $f(T(x))$. \zedit{To be consistent with the design of \oursys}, we apply E\&R operations 10 times for each image and calculate an arithmetic mean \zedit{of the 10 measurements}. The visualization of samples is shown in Figure~\ref{wd_kl}, which confirms our proposition; that is, the changes caused by E\&R operations on benign images and AEs fall into different clusters.

Next, we quantitatively analyse to what extent the distance/ divergence measurement 
can help discriminate an AE that is across the decision boundary. In detail, we use an optimal threshold based on the ROC (receiver operating characteristic) curve, to split AEs and benign images distributions. Table~\ref{tab:fpr_tpr} presents the FPR and TPR (i.e., Detection Rate defined in Section~\ref{sec:eval}). \zedit{Note that the results are only for illustrating that E\&R imposes different impacts
on AEs and benign samples in terms of} 
\fedit{probability mass function} 
\zedit{changes, and do not represent the detection
performance of $\oursys$ (see Section~\ref{sec:eval} for its detection performance). 
Here, we only}
\fedit{use} 
\zedit{one dimensional feature (i.e., the Wasserstein distance or KL divergence)}
\fedit{to split two clusters,}
\zedit{information loss inevitably degrades the}
\fedit{splitting} 
\zedit{performance, which is mitigated by the design of $\oursys$.}

\begin{figure*} 
\centering
\subfloat[Benign samples]{\includegraphics[scale=0.4]{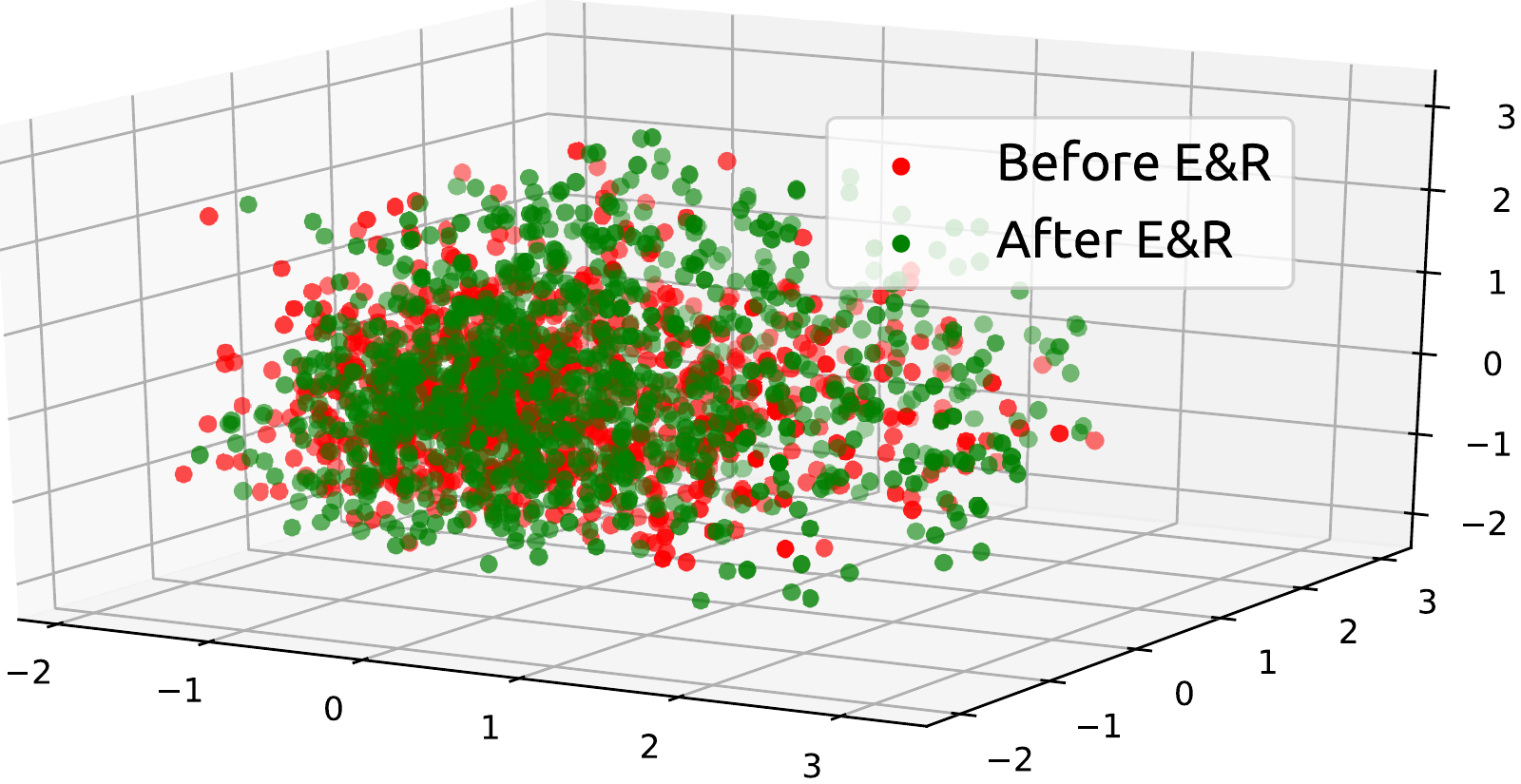}}
\ \ \ \ \ \ 
\subfloat[AEs]{\includegraphics[scale=0.4]{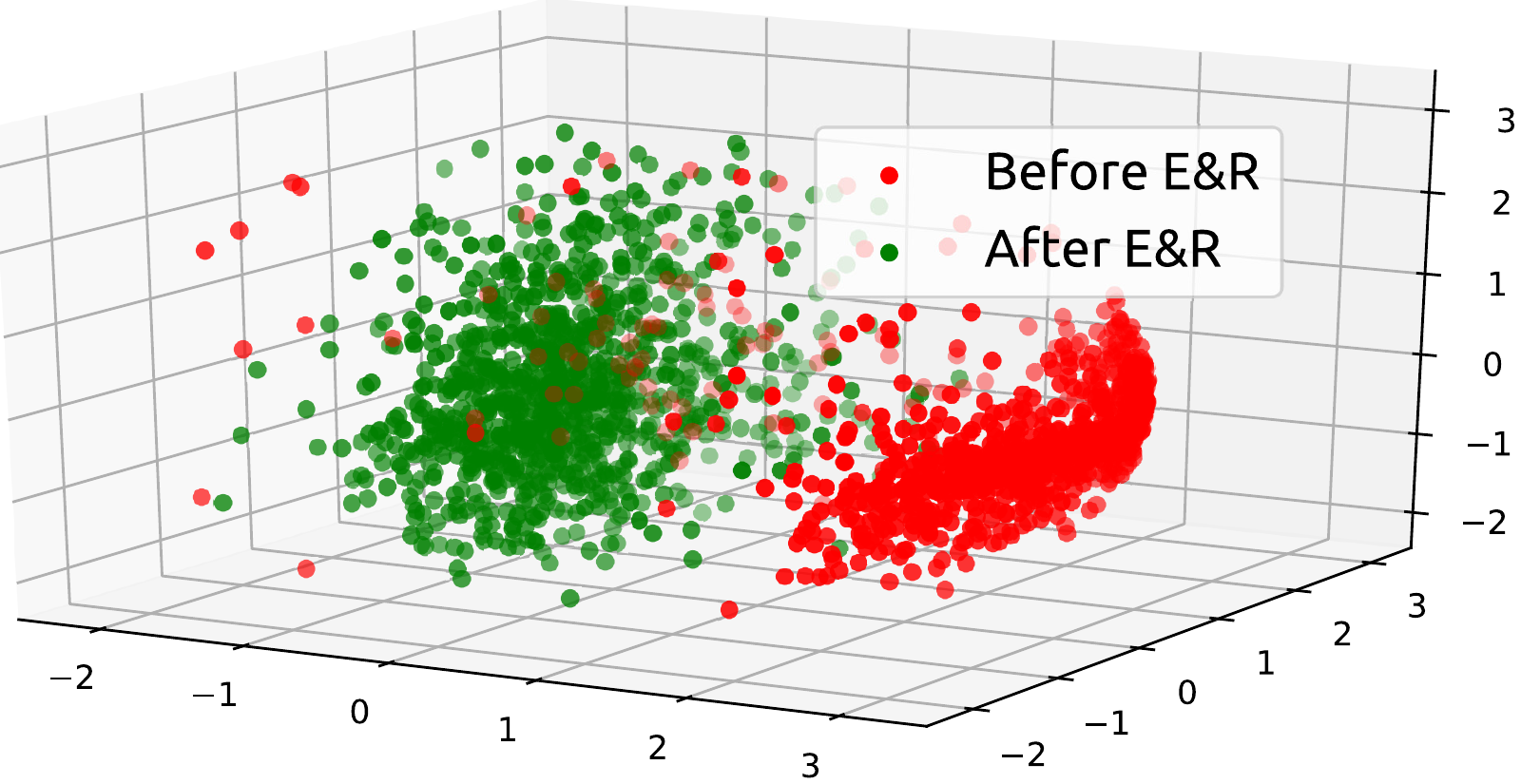}}
\af
\caption{Visualization of feature vectors. The coordinate axes respectively represent three largest principal components.}\label{fig_pca} 
\end{figure*}

\vspace{3pt}
\noindent \textbf{Interpretation through Visualization of Feature Vectors.}
The feature vectors due to 1,000 randomly selected benign samples from the
ImageNet dataset and the corresponding 1,000 AEs are visualized in Figure~\ref{fig_pca}.
For the visualization purpose, it shows only three principal components of the pre-processed feature vectors (see Figure~\ref{design}). We have two observations. (1) While the feature vectors of benign
samples, before and after the E\&R operations, are
close (Figure~\ref{fig_pca}(a)), 
those of AEs form two clusters
far apart (Figure~\ref{fig_pca}(b)). 
(2) PCA is effective in preserving features that help
distinguish benign samples from AEs.

\section{Related Work}\label{sec:relate}

Countermeasures against AE attacks can be roughly divided into two categories. The first category aims to eliminate the inﬂuences of AEs by either rectifying them or fortifying the target neural network itself. The second category is AE detectors (including our work), the goal of which is to predict whether an input is adversarial,
so that the target neural network can reject those inputs. Given the large body of research on AEs, this is not intended to be exhaustive.

\subsection{Adversarial Influences Elimination}

To improve the robustness of neural networks, \emph{adversarial training} augments the training set with the label-corrected AEs~\cite{zheng2016improving,madry2018towards}. Buckman et al.~\cite{buckman2018thermometer} propose using thermometer-encoded inputs to assist adversarial training. 
Alternatively, \emph{Shield}~\cite{das2018shield} enhances a model by re-training it with multiple levels of compressed images using JPEG, a commonly used image compression technique. 

Another strategy 
is to pre-process the inputs before feeding them to neural networks. For instance, the pixel deflection and a wavelet-based denoiser are combined to rectify AEs~\cite{prakash2018deflecting}. Liao et al.~\cite{liao2018defense} propose higher-level guided denoisers aiming to remove the adversarial noise from inputs. Some other methods adopt JPEG compression techniques 
~\cite{prakash2018protecting,guo2017countering} to filter out the information redundancy, which otherwise 
provides living space for adversarial perturbations. However, their accuracies under adaptive attacks are lack of adequate evaluations. CIIDefence~\cite{gupta2019ciidefence} proposes to use image inpainting with wavelet based denoising to rectify the classification result. However, its inpainting mask is guided by class activation maps, which can be predicted and exploited by an adaptive attacker. Both MagNet~\cite{meng2017magnet} and~\cite{chenposter} essentially take the path of removing noises/enhancing images, rather than the Erase-and-Restore path proposed in this work. REMIX~\cite{chenposter} applies inpainting to rectifying classification results, {with an rectifying accuracy
86\% on CIFAR-10}. It uses autoencoder as the inpainter. Autoencoders are typically data-specific, which means that it is only effective on images similar to what they have been trained on. {It did not
study the resilient to adaptive attacks and did not provide interpretation either.}

{Unlike all these works, the purpose of our work is for highly accurate attack detection, e.g., an accuracy of over 98\% 
on CIFAR-10 and ImageNet. It does not have dependency on high similarity
between training data and testing data. It is target-model agnostic: a detector trained using AEs targeting one model 
can be directly used to detect AEs targeting another.
Moreover, our work provides interpretation why the detection method works, and carefully examines its resilience to adaptive attacks.} 

\subsection{Adversarial Examples Detection}

Li et al.~\cite{li2017adversarial} extract PCA features 
after inner convolutional layers of the DNN, and then use
a cascade classifier to detect AEs. Metzen et al.~\cite{metzen2017detecting} train a CNN-based auxiliary network. This light-weight sub-network works with the target model to detect AEs. 
Some techniques apply pre-processors on input images and use prediction mismatch strategy to detect AEs. For example, Meng et al.~\cite{meng2017magnet} train an auto-encoder as the image filter. If the predictions of an original image and the corresponding processed one fail to match, the input is adversarial. Similarly, Xu et al.~\cite{xu2017feature} propose Feature Squeezing to detect AEs by comparing the prediction for the original input with that for the squeezed one. However, adaptive attacks have successfully circumvented all of the aforementioned detection methods~\cite{carlini2017adversarial,carlini2017magnet,he2017adversarial}.
Finally, Tian et al.~\cite{tian2018detecting} leverage image rotation and shifting as pre-processors to construct a detector. Although these operations can produce certain randomness to counter
some adaptive attacks, their randomness pool is very limited. It only has
45 possible transformations.
As a result, their method can merely achieve a detection rate of 70\% under adaptive attacks~\cite{tian2018detecting}. 

Zeng et al.~\cite{zeng2019multiversion} proposes a novel AE detection
method inspired by multiversion programming, which first uses multiple off-the-shelf audio 
recognition systems to classify the same audio input and then compares the classification
results to detect AEs. Their insight is the extraordinary difficulty of generating highly transferable audio AEs, which is not the case for image AEs. We also make use of multiple classification results, which,
however, is based on the idea of sampling (i.e., applying E\&R multiple times) to enhance the detection accuracy.

To our knowledge, our prior work~\cite{zuo2019l0} is the first that proposes to
use inpainting for AE detection, but it applies inpainting 
in a different way from this work. Specifically, ~\cite{zuo2019l0} focuses on detecting $L_0$ attacks by inpainting salient noises, as $L_0$ attacks
usually cause large-amplitude perturbations due to minimizing the number of modified pixels.

The AE detection idea that \emph{intentionally} and \emph{randomly} ``damages'' (i.e., erases) some pixels of an image and
then uses an inpainting algorithm  is not only ingenious and effective, but can also be interpreted and keep resilient to adaptive attacks. Unlike other very complex methods, our method
is extremely simple and easy to apply. As discussed in Section~\ref{sec:discuss}, although it only handles
$L_2$ attacks, it can easily work as a plugin or complement to enhance an existing attack detection system.

\section{Discussion and Future Work}\label{sec:discuss}

\begin{table}
\centering
\renewcommand\arraystretch{1.02}
\caption{Performance of integrating \oursys with an existing  detector~\cite{zuo2019l0}.}\label{tab:l0} 
\begin{tabular}{c||c|c|c|c|c}
\specialrule{.1em}{.05em}{.05em}
      \multirow{2}{*}{\textbf{Classifier}} & 
      \multirow{2}{*}{\textbf{FPR}} & \multicolumn{4}{c}{\textbf{Detection Rate}}  \\ \cline{3-6}
      
    &  & CW-$L_0$ &  JSMA & CW-$L_2$ & DeepFool\\ \specialrule{.1em}{.05em}{.05em}
{SVM} &  {3.4\%}   & {98.8}\% & {99.6\%} & {97.2\%} & {98.0\%} \\ \hline 
AdaBoost& 1.5\% & 98.8\% & 99.6\%   & 96.4\% & 97.2\% \\  
\specialrule{.1em}{.05em}{.05em}
\end{tabular}
\end{table}

While our work focuses on detecting $L_2$ AEs, it is easy to combine our approach with other detectors that
show strengths in detecting other types of AEs to build a comprehensive hybrid detector. A simplest integration is that \emph{an input is detected as an AE if any of the integrated detectors reports so}. 
To illustrate this, as an example, we integrate \oursys with our detection 
system~\cite{zuo2019l0} specialized in detecting $L_0$ attacks to build a more \emph{comprehensive} detector.
Table~\ref{tab:l0} shows the performance of this hybrid detector.

The proposed erasing and restoring approach works by 
destruction of the carefully perturbed pixels. 
Attackers thus may consider minimizing the number of perturbed pixels, like in $L_0$ AEs, 
to evade our detection. 
However, the prior work points out that $L_0$ AE generation
results in large amplitudes of altered pixels, which can be exploited to locate and restore
most of the maliciously perturbed pixels~\cite{zuo2019l0}. 
Therefore, for the purpose of AE generation, making a trade-off between the number of altered pixels
and their resulting amplitudes is a direction worth exploration. 

Another possible adaptive attack is to limit the perturbations in a restricted area that the defender is not aware of. Most prior works~\cite{shafahi2018adversarial, modas2019sparsefool, kwon2019restricted} that limit perturbed pixels to a given sub-region use $L_0$-norm. We notice that some recent works~\cite{dong2020greedyfool, deng2019generate} that only perturb pixels in a limited region also use $L_2$-norm to achieve better invisibility. However, their modified regions or even pixels are predictable, which can be exploited by an AE detector. Therefore, how to limit the $L_2$ perturbation to an \emph{arbitrary} sub-region is still an open question. A future task is to investigate
the effectiveness of E\&R once such $L_2$ perturbations are available.

This work focuses on attacks launched against digital images; we notice that physical attacks~\cite{eykholt2018robust, song2018physical} are attracting more and more interests from the research community. In particular, patch-based AEs, which are widely used in physical attacks, are not in the scope of this work. However, it is interesting to study the effectiveness of E\&R on  physical attacks~\cite{eykholt2018robust}. We leave this as our future work.

Finally, some recent studies on certified robustness have attracted much interest from the research community. For example, Cohen et al.~\cite{cohen2019certified} present 
a certified
robustness guarantee in $L_2$ norm for the smoothed classifier 
that is obtained by using Gaussian noise. Furthermore, Jia et al.~\cite{jia2019certified} derive a tight robustness in $L_2$ norm for top-$k$ predictions when using randomized smoothing with Gaussian noise. Some related works ~\cite{sanders2019inpainting, adam2017denoising} also show that inpainting has a side effect of denoising by smoothing the interpolated pixels. Our E\&R approach can be considered as an alternative to randomized smoothing. Thus, it is interesting to analyze the certified accuracy of our E\&R  method. We plan to explore this in our future work.

\vspace{-2pt}
\section{Conclusion}\label{sec:conclude}

Our finding has revealed 
that $L_2$ AEs are sensitive to 
the Erase-and-Restore
operations, while benign samples are not.
Exploiting the sensitivity distinction, we have proposed a novel and 
effective AE detection approach E\&R. It outperforms other state-of-the-art approaches in terms of 
both detection rates and false positive rates. In addition, our detector is target-model agnostic, keeps effective across different $L_2$ attack methods (i.e., good transferability across attack methods), and is resilient to adaptive attacks. 
Furthermore, we have interpreted the detection technique from both qualitative and quantitative angles to provide deeper understanding of the technique. {Unlike many
other detection methods that are complex and thus difficult to construct and train,
this method is very simple to build and easy to apply in practice.}




\begin{acks}

We would like to thank 
our shepherd, Dr. Qi Alfred Chen, and 
the anonymous reviewers for their invaluable suggestions. This work was supported in part 
by the US National Science Foundation (NSF) under grants CNS-1856380 and CNS-2016415.
\end{acks}

\bibliographystyle{ACM-Reference-Format}
\bibliography{sample-sigconf}


\begin{thebibliography}{55}


\ifx \showCODEN    \undefined \def \showCODEN     #1{\unskip}     \fi
\ifx \showDOI      \undefined \def \showDOI       #1{#1}\fi
\ifx \showISBNx    \undefined \def \showISBNx     #1{\unskip}     \fi
\ifx \showISBNxiii \undefined \def \showISBNxiii  #1{\unskip}     \fi
\ifx \showISSN     \undefined \def \showISSN      #1{\unskip}     \fi
\ifx \showLCCN     \undefined \def \showLCCN      #1{\unskip}     \fi
\ifx \shownote     \undefined \def \shownote      #1{#1}          \fi
\ifx \showarticletitle \undefined \def \showarticletitle #1{#1}   \fi
\ifx \showURL      \undefined \def \showURL       {\relax}        \fi
\providecommand\bibfield[2]{#2}
\providecommand\bibinfo[2]{#2}
\providecommand\natexlab[1]{#1}
\providecommand\showeprint[2][]{arXiv:#2}

\bibitem[\protect\citeauthoryear{Adam, Peter, and Weickert}{Adam
  et~al\mbox{.}}{2017}]%
        {adam2017denoising}
\bibfield{author}{\bibinfo{person}{Robin~Dirk Adam}, \bibinfo{person}{Pascal
  Peter}, {and} \bibinfo{person}{Joachim Weickert}.}
  \bibinfo{year}{2017}\natexlab{}.
\newblock \showarticletitle{Denoising by inpainting}. In
  \bibinfo{booktitle}{\emph{International Conference on Scale Space and
  Variational Methods in Computer Vision}}. Springer,
  \bibinfo{pages}{121--132}.
\newblock


\bibitem[\protect\citeauthoryear{Athalye, Carlini, and Wagner}{Athalye
  et~al\mbox{.}}{2018a}]%
        {obfuscated-gradients}
\bibfield{author}{\bibinfo{person}{Anish Athalye}, \bibinfo{person}{Nicholas
  Carlini}, {and} \bibinfo{person}{David Wagner}.}
  \bibinfo{year}{2018}\natexlab{a}.
\newblock \showarticletitle{Obfuscated Gradients Give a False Sense of
  Security: Circumventing Defenses to Adversarial Examples}. In
  \bibinfo{booktitle}{\emph{Proceedings of the 35th International Conference on
  Machine Learning (ICML)}}.
\newblock


\bibitem[\protect\citeauthoryear{Athalye, Engstrom, Ilyas, and Kwok}{Athalye
  et~al\mbox{.}}{2018b}]%
        {athalye2018synthesizing}
\bibfield{author}{\bibinfo{person}{Anish Athalye}, \bibinfo{person}{Logan
  Engstrom}, \bibinfo{person}{Andrew Ilyas}, {and} \bibinfo{person}{Kevin
  Kwok}.} \bibinfo{year}{2018}\natexlab{b}.
\newblock \showarticletitle{Synthesizing Robust Adversarial Examples}. In
  \bibinfo{booktitle}{\emph{Proceedings of the 35th International Conference on
  Machine Learning (ICML)}}.
\newblock


\bibitem[\protect\citeauthoryear{Buckman, Roy, Raffel, and Goodfellow}{Buckman
  et~al\mbox{.}}{2018}]%
        {buckman2018thermometer}
\bibfield{author}{\bibinfo{person}{Jacob Buckman}, \bibinfo{person}{Aurko Roy},
  \bibinfo{person}{Colin Raffel}, {and} \bibinfo{person}{Ian Goodfellow}.}
  \bibinfo{year}{2018}\natexlab{}.
\newblock \showarticletitle{Thermometer encoding: One hot way to resist
  adversarial examples}. In \bibinfo{booktitle}{\emph{International Conference
  on Learning Representations (ICLR)}}.
\newblock


\bibitem[\protect\citeauthoryear{Carlini and Wagner}{Carlini and
  Wagner}{2017a}]%
        {carlini2017adversarial}
\bibfield{author}{\bibinfo{person}{Nicholas Carlini} {and}
  \bibinfo{person}{David Wagner}.} \bibinfo{year}{2017}\natexlab{a}.
\newblock \showarticletitle{Adversarial examples are not easily detected:
  Bypassing ten detection methods}. In \bibinfo{booktitle}{\emph{ACM Workshop
  on Artificial Intelligence and Security}}.
\newblock


\bibitem[\protect\citeauthoryear{Carlini and Wagner}{Carlini and
  Wagner}{2017b}]%
        {carlini2017magnet}
\bibfield{author}{\bibinfo{person}{Nicholas Carlini} {and}
  \bibinfo{person}{David Wagner}.} \bibinfo{year}{2017}\natexlab{b}.
\newblock \showarticletitle{Magnet and ``efficient defenses against adversarial
  attacks" are not robust to adversarial examples}.
\newblock \bibinfo{journal}{\emph{arXiv preprint arXiv:1711.08478}}
  (\bibinfo{year}{2017}).
\newblock


\bibitem[\protect\citeauthoryear{Carlini and Wagner}{Carlini and
  Wagner}{2017c}]%
        {carlini2017towards}
\bibfield{author}{\bibinfo{person}{Nicholas Carlini} {and}
  \bibinfo{person}{David Wagner}.} \bibinfo{year}{2017}\natexlab{c}.
\newblock \showarticletitle{Towards evaluating the robustness of neural
  networks}. In \bibinfo{booktitle}{\emph{IEEE Symposium on Security and
  Privacy (SP)}}.
\newblock


\bibitem[\protect\citeauthoryear{Chen, Chen, and Yu}{Chen
  et~al\mbox{.}}{2018}]%
        {chenposter}
\bibfield{author}{\bibinfo{person}{Kang-Cheng Chen}, \bibinfo{person}{Pin-Yu
  Chen}, {and} \bibinfo{person}{Chia-Mu Yu}.} \bibinfo{year}{2018}\natexlab{}.
\newblock \showarticletitle{Poster: REMIX: Mitigating Adversarial Perturbation
  by Reforming, Masking and Inpainting}. In \bibinfo{booktitle}{\emph{IEEE
  Symposium on Security and Privacy (SP)}}.
\newblock


\bibitem[\protect\citeauthoryear{Chollet}{Chollet}{2015}]%
        {chollet2015keras}
\bibfield{author}{\bibinfo{person}{Fran\c{c}ois Chollet}.}
  \bibinfo{year}{2015}\natexlab{}.
\newblock \bibinfo{title}{Keras}.
\newblock \bibinfo{howpublished}{\url{https://keras.io}}.
\newblock


\bibitem[\protect\citeauthoryear{Cohen, Rosenfeld, and Kolter}{Cohen
  et~al\mbox{.}}{2019}]%
        {cohen2019certified}
\bibfield{author}{\bibinfo{person}{Jeremy~M Cohen}, \bibinfo{person}{Elan
  Rosenfeld}, {and} \bibinfo{person}{J~Zico Kolter}.}
  \bibinfo{year}{2019}\natexlab{}.
\newblock \showarticletitle{Certified adversarial robustness via randomized
  smoothing}. In \bibinfo{booktitle}{\emph{Proceedings of the 36th
  International Conference on Machine Learning (ICML)}}.
\newblock


\bibitem[\protect\citeauthoryear{Cortes and Vapnik}{Cortes and Vapnik}{1995}]%
        {cortes1995support}
\bibfield{author}{\bibinfo{person}{Corinna Cortes} {and}
  \bibinfo{person}{Vladimir Vapnik}.} \bibinfo{year}{1995}\natexlab{}.
\newblock \showarticletitle{Support-vector networks}.
\newblock \bibinfo{journal}{\emph{Machine learning}} \bibinfo{volume}{20},
  \bibinfo{number}{3} (\bibinfo{year}{1995}).
\newblock


\bibitem[\protect\citeauthoryear{da~Costa, Contato, Nazare, Neto, and
  Ponti}{da~Costa et~al\mbox{.}}{2016}]%
        {da2016empirical}
\bibfield{author}{\bibinfo{person}{Gabriel B~Paranhos da Costa},
  \bibinfo{person}{Welinton~A Contato}, \bibinfo{person}{Tiago~S Nazare},
  \bibinfo{person}{Jo{\~a}o~ES Neto}, {and} \bibinfo{person}{Moacir Ponti}.}
  \bibinfo{year}{2016}\natexlab{}.
\newblock \showarticletitle{An empirical study on the effects of different
  types of noise in image classification tasks}.
\newblock \bibinfo{journal}{\emph{arXiv preprint arXiv:1609.02781}}
  (\bibinfo{year}{2016}).
\newblock


\bibitem[\protect\citeauthoryear{Das, Shanbhogue, Chen, Hohman, Li, Chen,
  Kounavis, and Chau}{Das et~al\mbox{.}}{2018}]%
        {das2018shield}
\bibfield{author}{\bibinfo{person}{Nilaksh Das}, \bibinfo{person}{Madhuri
  Shanbhogue}, \bibinfo{person}{Shang-Tse Chen}, \bibinfo{person}{Fred Hohman},
  \bibinfo{person}{Siwei Li}, \bibinfo{person}{Li Chen},
  \bibinfo{person}{Michael~E Kounavis}, {and} \bibinfo{person}{Duen~Horng
  Chau}.} \bibinfo{year}{2018}\natexlab{}.
\newblock \showarticletitle{Shield: Fast, practical defense and vaccination for
  deep learning using {JPEG} compression}. In \bibinfo{booktitle}{\emph{ACM
  SIGKDD International Conference on Knowledge Discovery \& Data Mining}}.
\newblock


\bibitem[\protect\citeauthoryear{Deng and Zeng}{Deng and Zeng}{2019}]%
        {deng2019generate}
\bibfield{author}{\bibinfo{person}{Ting Deng} {and} \bibinfo{person}{Zhigang
  Zeng}.} \bibinfo{year}{2019}\natexlab{}.
\newblock \showarticletitle{Generate adversarial examples by spatially
  perturbing on the meaningful area}.
\newblock \bibinfo{journal}{\emph{Pattern Recognition Letters}}
  \bibinfo{volume}{125} (\bibinfo{year}{2019}), \bibinfo{pages}{632--638}.
\newblock


\bibitem[\protect\citeauthoryear{Diamond, Sitzmann, Boyd, Wetzstein, and
  Heide}{Diamond et~al\mbox{.}}{2017}]%
        {diamond2017dirty}
\bibfield{author}{\bibinfo{person}{Steven Diamond}, \bibinfo{person}{Vincent
  Sitzmann}, \bibinfo{person}{Stephen Boyd}, \bibinfo{person}{Gordon
  Wetzstein}, {and} \bibinfo{person}{Felix Heide}.}
  \bibinfo{year}{2017}\natexlab{}.
\newblock \showarticletitle{Dirty pixels: Optimizing image classification
  architectures for raw sensor data}.
\newblock \bibinfo{journal}{\emph{arXiv preprint arXiv:1701.06487}}
  (\bibinfo{year}{2017}).
\newblock


\bibitem[\protect\citeauthoryear{Dodge and Karam}{Dodge and Karam}{2016}]%
        {dodge2016understanding}
\bibfield{author}{\bibinfo{person}{Samuel Dodge} {and} \bibinfo{person}{Lina
  Karam}.} \bibinfo{year}{2016}\natexlab{}.
\newblock \showarticletitle{Understanding how image quality affects deep neural
  networks}. In \bibinfo{booktitle}{\emph{IEEE International Conference on
  Quality of Multimedia Experience (QoMEX)}}.
\newblock


\bibitem[\protect\citeauthoryear{Dong, Chen, Bao, Qin, Yuan, Zhang, Yu, and
  Chen}{Dong et~al\mbox{.}}{2020}]%
        {dong2020greedyfool}
\bibfield{author}{\bibinfo{person}{Xiaoyi Dong}, \bibinfo{person}{Dongdong
  Chen}, \bibinfo{person}{Jianmin Bao}, \bibinfo{person}{Chuan Qin},
  \bibinfo{person}{Lu Yuan}, \bibinfo{person}{Weiming Zhang},
  \bibinfo{person}{Nenghai Yu}, {and} \bibinfo{person}{Dong Chen}.}
  \bibinfo{year}{2020}\natexlab{}.
\newblock \showarticletitle{GreedyFool: Distortion-Aware Sparse Adversarial
  Attack}. In \bibinfo{booktitle}{\emph{Advances in Neural Information
  Processing Systems (NeurIPS)}}.
\newblock


\bibitem[\protect\citeauthoryear{Eykholt, Evtimov, Fernandes, Li, Rahmati,
  Tramer, Prakash, Kohno, and Song}{Eykholt et~al\mbox{.}}{2018a}]%
        {song2018physical}
\bibfield{author}{\bibinfo{person}{Kevin Eykholt}, \bibinfo{person}{Ivan
  Evtimov}, \bibinfo{person}{Earlence Fernandes}, \bibinfo{person}{Bo Li},
  \bibinfo{person}{Amir Rahmati}, \bibinfo{person}{Florian Tramer},
  \bibinfo{person}{Atul Prakash}, \bibinfo{person}{Tadayoshi Kohno}, {and}
  \bibinfo{person}{Dawn Song}.} \bibinfo{year}{2018}\natexlab{a}.
\newblock \showarticletitle{Physical adversarial examples for object
  detectors}. In \bibinfo{booktitle}{\emph{12th {USENIX} Workshop on Offensive
  Technologies ({WOOT})}}.
\newblock


\bibitem[\protect\citeauthoryear{Eykholt, Evtimov, Fernandes, Li, Rahmati,
  Xiao, Prakash, Kohno, and Song}{Eykholt et~al\mbox{.}}{2018b}]%
        {eykholt2018robust}
\bibfield{author}{\bibinfo{person}{Kevin Eykholt}, \bibinfo{person}{Ivan
  Evtimov}, \bibinfo{person}{Earlence Fernandes}, \bibinfo{person}{Bo Li},
  \bibinfo{person}{Amir Rahmati}, \bibinfo{person}{Chaowei Xiao},
  \bibinfo{person}{Atul Prakash}, \bibinfo{person}{Tadayoshi Kohno}, {and}
  \bibinfo{person}{Dawn Song}.} \bibinfo{year}{2018}\natexlab{b}.
\newblock \showarticletitle{Robust physical-world attacks on deep learning
  visual classification}. In \bibinfo{booktitle}{\emph{Conference on Computer
  Vision and Pattern Recognition (CVPR)}}.
\newblock


\bibitem[\protect\citeauthoryear{Freund and Schapire}{Freund and
  Schapire}{1997}]%
        {freund1997decision}
\bibfield{author}{\bibinfo{person}{Yoav Freund} {and} \bibinfo{person}{Robert~E
  Schapire}.} \bibinfo{year}{1997}\natexlab{}.
\newblock \showarticletitle{A decision-theoretic generalization of on-line
  learning and an application to boosting}.
\newblock \bibinfo{journal}{\emph{Journal of computer and system sciences}}
  \bibinfo{volume}{55}, \bibinfo{number}{1} (\bibinfo{year}{1997}).
\newblock


\bibitem[\protect\citeauthoryear{Gong, Wang, and Ku}{Gong
  et~al\mbox{.}}{2017}]%
        {gong2017adversarial}
\bibfield{author}{\bibinfo{person}{Zhitao Gong}, \bibinfo{person}{Wenlu Wang},
  {and} \bibinfo{person}{Wei-Shinn Ku}.} \bibinfo{year}{2017}\natexlab{}.
\newblock \showarticletitle{Adversarial and clean data are not twins}.
\newblock \bibinfo{journal}{\emph{arXiv preprint arXiv:1704.04960}}
  (\bibinfo{year}{2017}).
\newblock


\bibitem[\protect\citeauthoryear{Grosse, Manoharan, Papernot, Backes, and
  McDaniel}{Grosse et~al\mbox{.}}{2017}]%
        {grosse2017statistical}
\bibfield{author}{\bibinfo{person}{Kathrin Grosse}, \bibinfo{person}{Praveen
  Manoharan}, \bibinfo{person}{Nicolas Papernot}, \bibinfo{person}{Michael
  Backes}, {and} \bibinfo{person}{Patrick McDaniel}.}
  \bibinfo{year}{2017}\natexlab{}.
\newblock \showarticletitle{On the (statistical) detection of adversarial
  examples}.
\newblock \bibinfo{journal}{\emph{arXiv preprint arXiv:1702.06280}}
  (\bibinfo{year}{2017}).
\newblock


\bibitem[\protect\citeauthoryear{Guo, Rana, Cisse, and van~der Maaten}{Guo
  et~al\mbox{.}}{2018}]%
        {guo2017countering}
\bibfield{author}{\bibinfo{person}{Chuan Guo}, \bibinfo{person}{Mayank Rana},
  \bibinfo{person}{Moustapha Cisse}, {and} \bibinfo{person}{Laurens van~der
  Maaten}.} \bibinfo{year}{2018}\natexlab{}.
\newblock \showarticletitle{Countering adversarial images using input
  transformations}. In \bibinfo{booktitle}{\emph{International Conference on
  Learning Representations (ICLR)}}.
\newblock


\bibitem[\protect\citeauthoryear{Gupta and Rahtu}{Gupta and Rahtu}{2019}]%
        {gupta2019ciidefence}
\bibfield{author}{\bibinfo{person}{Puneet Gupta} {and} \bibinfo{person}{Esa
  Rahtu}.} \bibinfo{year}{2019}\natexlab{}.
\newblock \showarticletitle{CIIDefence: Defeating Adversarial Attacks by Fusing
  Class-Specific Image Inpainting and Image Denoising}. In
  \bibinfo{booktitle}{\emph{Proceedings of the IEEE International Conference on
  Computer Vision (ICCV)}}.
\newblock


\bibitem[\protect\citeauthoryear{He, Zhang, Ren, and Sun}{He
  et~al\mbox{.}}{2016}]%
        {he2016deep}
\bibfield{author}{\bibinfo{person}{Kaiming He}, \bibinfo{person}{Xiangyu
  Zhang}, \bibinfo{person}{Shaoqing Ren}, {and} \bibinfo{person}{Jian Sun}.}
  \bibinfo{year}{2016}\natexlab{}.
\newblock \showarticletitle{Deep residual learning for image recognition}. In
  \bibinfo{booktitle}{\emph{Conference on Computer Vision and Pattern
  Recognition (CVPR)}}.
\newblock


\bibitem[\protect\citeauthoryear{He, Wei, Chen, Carlini, and Song}{He
  et~al\mbox{.}}{2017}]%
        {he2017adversarial}
\bibfield{author}{\bibinfo{person}{Warren He}, \bibinfo{person}{James Wei},
  \bibinfo{person}{Xinyun Chen}, \bibinfo{person}{Nicholas Carlini}, {and}
  \bibinfo{person}{Dawn Song}.} \bibinfo{year}{2017}\natexlab{}.
\newblock \showarticletitle{Adversarial example defenses: Ensembles of weak
  defenses are not strong}. In \bibinfo{booktitle}{\emph{11th {USENIX} Workshop
  on Offensive Technologies (WOOT)}}.
\newblock


\bibitem[\protect\citeauthoryear{Jia, Cao, Wang, and Gong}{Jia
  et~al\mbox{.}}{2020}]%
        {jia2019certified}
\bibfield{author}{\bibinfo{person}{Jinyuan Jia}, \bibinfo{person}{Xiaoyu Cao},
  \bibinfo{person}{Binghui Wang}, {and} \bibinfo{person}{Neil~Zhenqiang Gong}.}
  \bibinfo{year}{2020}\natexlab{}.
\newblock \showarticletitle{Certified robustness for top-k predictions against
  adversarial perturbations via randomized smoothing}. In
  \bibinfo{booktitle}{\emph{International Conference on Learning
  Representations (ICLR)}}.
\newblock


\bibitem[\protect\citeauthoryear{Kingma and Ba}{Kingma and Ba}{2015}]%
        {kingma2014adam}
\bibfield{author}{\bibinfo{person}{Diederik~P Kingma} {and}
  \bibinfo{person}{Jimmy Ba}.} \bibinfo{year}{2015}\natexlab{}.
\newblock \showarticletitle{Adam: A method for stochastic optimization}. In
  \bibinfo{booktitle}{\emph{International Conference on Learning
  Representations (ICLR)}}.
\newblock


\bibitem[\protect\citeauthoryear{Kullback}{Kullback}{1997}]%
        {kullback1997information}
\bibfield{author}{\bibinfo{person}{Solomon Kullback}.}
  \bibinfo{year}{1997}\natexlab{}.
\newblock \bibinfo{booktitle}{\emph{Information theory and statistics}}.
\newblock \bibinfo{publisher}{Courier Corporation}.
\newblock


\bibitem[\protect\citeauthoryear{Kwon, Yoon, and Choi}{Kwon
  et~al\mbox{.}}{2019}]%
        {kwon2019restricted}
\bibfield{author}{\bibinfo{person}{Hyun Kwon}, \bibinfo{person}{Hyunsoo Yoon},
  {and} \bibinfo{person}{Daeseon Choi}.} \bibinfo{year}{2019}\natexlab{}.
\newblock \showarticletitle{Restricted evasion attack: Generation of
  restricted-area adversarial example}.
\newblock \bibinfo{journal}{\emph{IEEE Access}}  \bibinfo{volume}{7}
  (\bibinfo{year}{2019}), \bibinfo{pages}{60908--60919}.
\newblock


\bibitem[\protect\citeauthoryear{Li and Li}{Li and Li}{2017}]%
        {li2017adversarial}
\bibfield{author}{\bibinfo{person}{Xin Li} {and} \bibinfo{person}{Fuxin Li}.}
  \bibinfo{year}{2017}\natexlab{}.
\newblock \showarticletitle{Adversarial examples detection in deep networks
  with convolutional filter statistics}. In
  \bibinfo{booktitle}{\emph{Proceedings of the IEEE International Conference on
  Computer Vision (ICCV)}}.
\newblock


\bibitem[\protect\citeauthoryear{Liao, Liang, Dong, Pang, Zhu, and Hu}{Liao
  et~al\mbox{.}}{2018}]%
        {liao2018defense}
\bibfield{author}{\bibinfo{person}{Fangzhou Liao}, \bibinfo{person}{Ming
  Liang}, \bibinfo{person}{Yinpeng Dong}, \bibinfo{person}{Tianyu Pang},
  \bibinfo{person}{Jun Zhu}, {and} \bibinfo{person}{Xiaolin Hu}.}
  \bibinfo{year}{2018}\natexlab{}.
\newblock \showarticletitle{Defense against adversarial attacks using
  high-level representation guided denoiser}. In
  \bibinfo{booktitle}{\emph{Conference on Computer Vision and Pattern
  Recognition (CVPR)}}.
\newblock


\bibitem[\protect\citeauthoryear{Ma, Liu, Tao, Lee, and Zhang}{Ma
  et~al\mbox{.}}{2019}]%
        {ma2019nic}
\bibfield{author}{\bibinfo{person}{Shiqing Ma}, \bibinfo{person}{Yingqi Liu},
  \bibinfo{person}{Guanhong Tao}, \bibinfo{person}{Wen-Chuan Lee}, {and}
  \bibinfo{person}{Xiangyu Zhang}.} \bibinfo{year}{2019}\natexlab{}.
\newblock \showarticletitle{{NIC}: Detecting Adversarial Samples with Neural
  Network Invariant Checking}. In \bibinfo{booktitle}{\emph{Network and
  Distributed System Security Symposium (NDSS)}}.
\newblock


\bibitem[\protect\citeauthoryear{Ma, Li, Wang, Erfani, Wijewickrema,
  Schoenebeck, Song, Houle, and Bailey}{Ma et~al\mbox{.}}{2018}]%
        {ma2018characterizing}
\bibfield{author}{\bibinfo{person}{Xingjun Ma}, \bibinfo{person}{Bo Li},
  \bibinfo{person}{Yisen Wang}, \bibinfo{person}{Sarah~M Erfani},
  \bibinfo{person}{Sudanthi Wijewickrema}, \bibinfo{person}{Grant Schoenebeck},
  \bibinfo{person}{Dawn Song}, \bibinfo{person}{Michael~E Houle}, {and}
  \bibinfo{person}{James Bailey}.} \bibinfo{year}{2018}\natexlab{}.
\newblock \showarticletitle{Characterizing adversarial subspaces using local
  intrinsic dimensionality}. In \bibinfo{booktitle}{\emph{International
  Conference on Learning Representations (ICLR)}}.
\newblock


\bibitem[\protect\citeauthoryear{Madry, Makelov, Schmidt, Tsipras, and
  Vladu}{Madry et~al\mbox{.}}{2018}]%
        {madry2018towards}
\bibfield{author}{\bibinfo{person}{Aleksander Madry},
  \bibinfo{person}{Aleksandar Makelov}, \bibinfo{person}{Ludwig Schmidt},
  \bibinfo{person}{Dimitris Tsipras}, {and} \bibinfo{person}{Adrian Vladu}.}
  \bibinfo{year}{2018}\natexlab{}.
\newblock \showarticletitle{Towards deep learning models resistant to
  adversarial attacks}. In \bibinfo{booktitle}{\emph{International Conference
  on Learning Representations (ICLR)}}.
\newblock


\bibitem[\protect\citeauthoryear{Mairal, Elad, and Sapiro}{Mairal
  et~al\mbox{.}}{2007}]%
        {mairal2007sparse}
\bibfield{author}{\bibinfo{person}{Julien Mairal}, \bibinfo{person}{Michael
  Elad}, {and} \bibinfo{person}{Guillermo Sapiro}.}
  \bibinfo{year}{2007}\natexlab{}.
\newblock \showarticletitle{Sparse representation for color image restoration}.
\newblock \bibinfo{journal}{\emph{IEEE Transactions on image processing}}
  \bibinfo{volume}{17}, \bibinfo{number}{1} (\bibinfo{year}{2007}).
\newblock


\bibitem[\protect\citeauthoryear{Meng and Chen}{Meng and Chen}{2017}]%
        {meng2017magnet}
\bibfield{author}{\bibinfo{person}{Dongyu Meng} {and} \bibinfo{person}{Hao
  Chen}.} \bibinfo{year}{2017}\natexlab{}.
\newblock \showarticletitle{Mag{N}et: a two-pronged defense against adversarial
  examples}. In \bibinfo{booktitle}{\emph{ACM SIGSAC Conference on Computer and
  Communications Security (CCS)}}.
\newblock


\bibitem[\protect\citeauthoryear{Metzen, Genewein, Fischer, and
  Bischoff}{Metzen et~al\mbox{.}}{2017}]%
        {metzen2017detecting}
\bibfield{author}{\bibinfo{person}{Jan~Hendrik Metzen}, \bibinfo{person}{Tim
  Genewein}, \bibinfo{person}{Volker Fischer}, {and} \bibinfo{person}{Bastian
  Bischoff}.} \bibinfo{year}{2017}\natexlab{}.
\newblock \showarticletitle{On detecting adversarial perturbations}. In
  \bibinfo{booktitle}{\emph{International Conference on Learning
  Representations (ICLR)}}.
\newblock


\bibitem[\protect\citeauthoryear{Modas, Moosavi-Dezfooli, and Frossard}{Modas
  et~al\mbox{.}}{2019}]%
        {modas2019sparsefool}
\bibfield{author}{\bibinfo{person}{Apostolos Modas},
  \bibinfo{person}{Seyed-Mohsen Moosavi-Dezfooli}, {and}
  \bibinfo{person}{Pascal Frossard}.} \bibinfo{year}{2019}\natexlab{}.
\newblock \showarticletitle{Sparsefool: a few pixels make a big difference}. In
  \bibinfo{booktitle}{\emph{Conference on Computer Vision and Pattern
  Recognition (CVPR)}}.
\newblock


\bibitem[\protect\citeauthoryear{Moosavi-Dezfooli, Fawzi, and
  Frossard}{Moosavi-Dezfooli et~al\mbox{.}}{2016}]%
        {moosavi2016deepfool}
\bibfield{author}{\bibinfo{person}{Seyed-Mohsen Moosavi-Dezfooli},
  \bibinfo{person}{Alhussein Fawzi}, {and} \bibinfo{person}{Pascal Frossard}.}
  \bibinfo{year}{2016}\natexlab{}.
\newblock \showarticletitle{Deep{F}ool: a simple and accurate method to fool
  deep neural networks}. In \bibinfo{booktitle}{\emph{Conference on Computer
  Vision and Pattern Recognition (CVPR)}}.
\newblock


\bibitem[\protect\citeauthoryear{Nicolae, Sinn, Tran, Buesser, Rawat, Wistuba,
  Zantedeschi, Baracaldo, Chen, Ludwig, Molloy, and Edwards}{Nicolae
  et~al\mbox{.}}{2018}]%
        {art2018}
\bibfield{author}{\bibinfo{person}{Maria-Irina Nicolae},
  \bibinfo{person}{Mathieu Sinn}, \bibinfo{person}{Minh~Ngoc Tran},
  \bibinfo{person}{Beat Buesser}, \bibinfo{person}{Ambrish Rawat},
  \bibinfo{person}{Martin Wistuba}, \bibinfo{person}{Valentina Zantedeschi},
  \bibinfo{person}{Nathalie Baracaldo}, \bibinfo{person}{Bryant Chen},
  \bibinfo{person}{Heiko Ludwig}, \bibinfo{person}{Ian Molloy}, {and}
  \bibinfo{person}{Ben Edwards}.} \bibinfo{year}{2018}\natexlab{}.
\newblock \bibinfo{title}{Adversarial Robustness Toolbox v1.0.1}.
\newblock
  \bibinfo{howpublished}{\url{https://adversarial-robustness-toolbox.readthedocs.io/en/latest/modules/attacks/evasion.html\#carlini-and-wagner-l-2-attack}}.
\newblock


\bibitem[\protect\citeauthoryear{Papernot, McDaniel, Wu, Jha, and
  Swami}{Papernot et~al\mbox{.}}{2016}]%
        {papernot2015distillation}
\bibfield{author}{\bibinfo{person}{Nicolas Papernot}, \bibinfo{person}{Patrick
  McDaniel}, \bibinfo{person}{Xi Wu}, \bibinfo{person}{Somesh Jha}, {and}
  \bibinfo{person}{Ananthram Swami}.} \bibinfo{year}{2016}\natexlab{}.
\newblock \showarticletitle{Distillation as a defense to adversarial
  perturbations against deep neural networks}. In
  \bibinfo{booktitle}{\emph{IEEE Symposium on Security and Privacy (SP)}}.
\newblock


\bibitem[\protect\citeauthoryear{Prakash, Moran, Garber, DiLillo, and
  Storer}{Prakash et~al\mbox{.}}{2018a}]%
        {prakash2018deflecting}
\bibfield{author}{\bibinfo{person}{Aaditya Prakash}, \bibinfo{person}{Nick
  Moran}, \bibinfo{person}{Solomon Garber}, \bibinfo{person}{Antonella
  DiLillo}, {and} \bibinfo{person}{James Storer}.}
  \bibinfo{year}{2018}\natexlab{a}.
\newblock \showarticletitle{Deflecting adversarial attacks with pixel
  deflection}. In \bibinfo{booktitle}{\emph{Conference on Computer Vision and
  Pattern Recognition (CVPR)}}.
\newblock


\bibitem[\protect\citeauthoryear{Prakash, Moran, Garber, DiLillo, and
  Storer}{Prakash et~al\mbox{.}}{2018b}]%
        {prakash2018protecting}
\bibfield{author}{\bibinfo{person}{Aaditya Prakash}, \bibinfo{person}{Nick
  Moran}, \bibinfo{person}{Solomon Garber}, \bibinfo{person}{Antonella
  DiLillo}, {and} \bibinfo{person}{James Storer}.}
  \bibinfo{year}{2018}\natexlab{b}.
\newblock \showarticletitle{Protecting JPEG images against adversarial
  attacks}. In \bibinfo{booktitle}{\emph{2018 Data Compression Conference}}.
  IEEE, \bibinfo{pages}{137--146}.
\newblock


\bibitem[\protect\citeauthoryear{Rauber, Brendel, and Bethge}{Rauber
  et~al\mbox{.}}{2017}]%
        {rauber2017foolbox}
\bibfield{author}{\bibinfo{person}{Jonas Rauber}, \bibinfo{person}{Wieland
  Brendel}, {and} \bibinfo{person}{Matthias Bethge}.}
  \bibinfo{year}{2017}\natexlab{}.
\newblock \showarticletitle{Foolbox: A {P}ython toolbox to benchmark the
  robustness of machine learning models}.
\newblock \bibinfo{journal}{\emph{arXiv preprint arXiv:1707.04131}}
  (\bibinfo{year}{2017}).
\newblock


\bibitem[\protect\citeauthoryear{Sanders and Dwyer}{Sanders and Dwyer}{2019}]%
        {sanders2019inpainting}
\bibfield{author}{\bibinfo{person}{Toby Sanders} {and}
  \bibinfo{person}{Christian Dwyer}.} \bibinfo{year}{2019}\natexlab{}.
\newblock \showarticletitle{Inpainting versus denoising for dose reduction in
  scanning-beam microscopies}.
\newblock \bibinfo{journal}{\emph{IEEE Transactions on Image Processing}}
  \bibinfo{volume}{29} (\bibinfo{year}{2019}), \bibinfo{pages}{351--359}.
\newblock


\bibitem[\protect\citeauthoryear{Shafahi, Huang, Studer, Feizi, and
  Goldstein}{Shafahi et~al\mbox{.}}{2019}]%
        {shafahi2018adversarial}
\bibfield{author}{\bibinfo{person}{Ali Shafahi}, \bibinfo{person}{W~Ronny
  Huang}, \bibinfo{person}{Christoph Studer}, \bibinfo{person}{Soheil Feizi},
  {and} \bibinfo{person}{Tom Goldstein}.} \bibinfo{year}{2019}\natexlab{}.
\newblock \showarticletitle{Are adversarial examples inevitable?}. In
  \bibinfo{booktitle}{\emph{International Conference on Learning
  Representations (ICLR)}}.
\newblock


\bibitem[\protect\citeauthoryear{Shen and Chan}{Shen and Chan}{2002}]%
        {shen2002mathematical}
\bibfield{author}{\bibinfo{person}{Jianhong Shen} {and} \bibinfo{person}{Tony~F
  Chan}.} \bibinfo{year}{2002}\natexlab{}.
\newblock \showarticletitle{Mathematical models for local nontexture
  inpaintings}.
\newblock \bibinfo{journal}{\emph{SIAM J. Appl. Math.}} \bibinfo{volume}{62},
  \bibinfo{number}{3} (\bibinfo{year}{2002}).
\newblock


\bibitem[\protect\citeauthoryear{Telea}{Telea}{2004}]%
        {telea2004image}
\bibfield{author}{\bibinfo{person}{Alexandru Telea}.}
  \bibinfo{year}{2004}\natexlab{}.
\newblock \showarticletitle{An image inpainting technique based on the fast
  marching method}.
\newblock \bibinfo{journal}{\emph{Journal of Graphics Tools}}
  \bibinfo{volume}{9}, \bibinfo{number}{1} (\bibinfo{year}{2004}).
\newblock


\bibitem[\protect\citeauthoryear{Tian, Yang, and Cai}{Tian
  et~al\mbox{.}}{2018}]%
        {tian2018detecting}
\bibfield{author}{\bibinfo{person}{Shixin Tian}, \bibinfo{person}{Guolei Yang},
  {and} \bibinfo{person}{Ying Cai}.} \bibinfo{year}{2018}\natexlab{}.
\newblock \showarticletitle{Detecting Adversarial Examples through Image
  Transformation}. In \bibinfo{booktitle}{\emph{AAAI Conference on Artificial
  Intelligence}}.
\newblock


\bibitem[\protect\citeauthoryear{Villani}{Villani}{2009}]%
        {villani2009wasserstein}
\bibfield{author}{\bibinfo{person}{C{\'e}dric Villani}.}
  \bibinfo{year}{2009}\natexlab{}.
\newblock \showarticletitle{The {W}asserstein distances}.
\newblock In \bibinfo{booktitle}{\emph{Optimal Transport}}.
  \bibinfo{publisher}{Springer}, \bibinfo{pages}{93--111}.
\newblock


\bibitem[\protect\citeauthoryear{Xu, Evans, and Qi}{Xu et~al\mbox{.}}{2018}]%
        {xu2017feature}
\bibfield{author}{\bibinfo{person}{Weilin Xu}, \bibinfo{person}{David Evans},
  {and} \bibinfo{person}{Yanjun Qi}.} \bibinfo{year}{2018}\natexlab{}.
\newblock \showarticletitle{Feature squeezing: Detecting adversarial examples
  in deep neural networks}. In \bibinfo{booktitle}{\emph{Network and
  Distributed System Security Symposium (NDSS)}}.
\newblock


\bibitem[\protect\citeauthoryear{Zeng, Su, Fu, Kayas, Luo, Du, Tan, and
  Wu}{Zeng et~al\mbox{.}}{2019}]%
        {zeng2019multiversion}
\bibfield{author}{\bibinfo{person}{Qiang Zeng}, \bibinfo{person}{Jianhai Su},
  \bibinfo{person}{Chenglong Fu}, \bibinfo{person}{Golam Kayas},
  \bibinfo{person}{Lannan Luo}, \bibinfo{person}{Xiaojiang Du},
  \bibinfo{person}{Chiu~C Tan}, {and} \bibinfo{person}{Jie Wu}.}
  \bibinfo{year}{2019}\natexlab{}.
\newblock \showarticletitle{A multiversion programming inspired approach to
  detecting audio adversarial examples}. In \bibinfo{booktitle}{\emph{49th
  Annual IEEE/IFIP International Conference on Dependable Systems and Networks
  (DSN)}}.
\newblock


\bibitem[\protect\citeauthoryear{Zheng, Song, Leung, and Goodfellow}{Zheng
  et~al\mbox{.}}{2016}]%
        {zheng2016improving}
\bibfield{author}{\bibinfo{person}{Stephan Zheng}, \bibinfo{person}{Yang Song},
  \bibinfo{person}{Thomas Leung}, {and} \bibinfo{person}{Ian Goodfellow}.}
  \bibinfo{year}{2016}\natexlab{}.
\newblock \showarticletitle{Improving the robustness of deep neural networks
  via stability training}. In \bibinfo{booktitle}{\emph{Conference on Computer
  Vision and Pattern Recognition (CVPR)}}.
\newblock


\bibitem[\protect\citeauthoryear{Zuo, Yang, Li, Luo, and Zeng}{Zuo
  et~al\mbox{.}}{2019}]%
        {zuo2019l0}
\bibfield{author}{\bibinfo{person}{Fei Zuo}, \bibinfo{person}{Bokai Yang},
  \bibinfo{person}{Xiaopeng Li}, \bibinfo{person}{Lannan Luo}, {and}
  \bibinfo{person}{Qiang Zeng}.} \bibinfo{year}{2019}\natexlab{}.
\newblock \showarticletitle{Exploiting the Inherent Limitation of ${L}_0$
  Adversarial Examples}. In \bibinfo{booktitle}{\emph{International Symposium
  on Research in Attacks, Intrusions and Defenses (RAID)}}.
\newblock


\end{thebibliography}

\end{document}